\documentclass[12]{amsart}
\usepackage{amssymb}
\usepackage{amsmath}
\usepackage{bbold}
\usepackage{bm}
\usepackage{braket}
\usepackage[title]{appendix}
\usepackage[dvipsnames]{xcolor}
\usepackage{soul}
\usepackage{graphicx}
\usepackage{enumerate}
\usepackage{verbatim}
\usepackage{enumitem}
\usepackage[english]{babel} 
\usepackage{natbib}

\usepackage{hyperref}
\hypersetup{
    colorlinks,
     linkcolor={red!80!black},
     citecolor={green!70!black},
     urlcolor={blue!80!black}
}
\usepackage{lipsum, babel}
\graphicspath{{../figura1/}{../figura2/}}
\usepackage{graphicx}

\usepackage{bbold}
\usepackage{bm}
\usepackage{a4wide}

\usepackage{soul}

\newcommand{\RE}{\mathbb{R}}
\newcommand{\N}{\mathbb{N}}

\newcommand{\bx}{\mathbf{x}}
\newcommand{\by}{\mathbf{y}}
\newcommand{\bs}{\mathbf{s}}
\newcommand{\CS}{\mathcal{S}}
\newcommand{\CF}{\mathcal{F}}
\newcommand{\CG}{\mathcal{G}}
\newcommand{\CH}{\mathcal{H}}

\newcommand{\bZ}{\mathbf{Z}}
\newcommand{\bz}{\mathbf{z}}

\newcommand{\br}{\mathbf{r}}

\newcommand{\bnn}{\mathbf{n}}

\newcommand{\bX}{\mathbf{X}}

\newcommand{\bY}{\mathbf{Y}}

\newcommand{\CN}{\mathcal{N}}
\newcommand{\CR}{\mathcal{R}}

\newcommand{\E}{\mathbf{E}}
\newcommand{\Var}{\mathbf{Var}}

\DeclareMathOperator{\tr}{tr}
\DeclareMathOperator{\vvec}{vec}
\DeclareMathOperator{\Cov}{Cov}

\usepackage{xcite}
\usepackage{xr}
\newcommand{\bS}{\mathbf{S}}
\newcommand{\CL}{\mathcal{L}}
\newcommand{\eI}{\mathbb{1}}

\newtheorem{definition}{Definizione}[section]
\newtheorem{lemma}[definition]{Lemma}
\newtheorem{proposition}[definition]{Proposition}

\newtheorem{remark}{Remark}

\title[Deep linear neural network in the proportional limit]{Proportional infinite-width infinite-depth limit  for deep lInear neural networks} 
\author{ Federico Bassetti$^*$,  Lucia Ladelli$^\dag$, Pietro Rotondo$^\ddagger$ }
\thanks{$^*$ email: federico.bassetti@polimi.it; address:  Dipartimento di Matematica,  Politecnico di Milano,  Milano, Italy.}
\thanks{$^\dag$  email: lucia.ladelli@polimi.it;  address: Dipartimento di Matematica,  Politecnico di Milano,  Milano, Italy.}
\thanks{$^\ddag $  email: pietro.rotondo@unipr.it;  address: Dipartimento di Scienze Matematiche, Fisiche e Informatiche\\
       Università degli Studi di Parma, Parma, Italy.}

\begin{document}

\maketitle

\begin{abstract}
We study the distributional properties of linear neural networks with random parameters in the context of large networks, where  the number of layers
 diverges in proportion to the number of neurons  per layer. 
  Prior works have shown that in the infinite-width regime, where the number of neurons per layer grows to infinity while the depth remains fixed, neural networks converge to a Gaussian process, known as the Neural Network Gaussian Process. However, this Gaussian limit sacrifices descriptive power, 
  as it lacks the ability to learn dependent features and produce output correlations that reflect observed labels. Motivated by these limitations, we explore the joint proportional limit in which both depth and width diverge but maintain a constant ratio, yielding a non-Gaussian distribution that retains correlations between outputs. Our contribution extends previous works by rigorously characterizing, for linear activation functions, the limiting distribution  as a nontrivial mixture of Gaussians. 
\end{abstract}
\section{Introduction}
A neural network can be viewed as a parameterized function \( f(\cdot|{\theta}) : \mathbb{R}^{N_0} \to \mathbb{R}^{D} \), 
recursively defined by linear transformations composed with non-linear activation functions. When a network architecture is fixed, parameters are learned using a training dataset. In this context  understanding the behaviour of neural networks with randomly initialized parameters is essential. Indeed, a randomly initialized network can be viewed as either the starting point for optimization or as the prior in a Bayesian learning framework.

In large neural networks with many neurons per layer, key insights arise through scaling limits, in particular the infinite-width limit, where the depth is fixed, and the width (number of neurons) grows to infinity. 
In this context, gaussian universality appears  both in training under gradient flow, described by the neural tangent kernel  \cite{JacotNTK}, and in the Bayesian inference setting, where exact relations between neural networks and kernel methods have been established \cite{lee2018deep,g.2018gaussian}.

Neal's seminal work \cite{Neal} established that Bayesian shallow networks (networks with one hidden layer and large width) converge to a Gaussian process under standard assumptions on the weight distribution. Following this result, research has been extended to multi-layer networks with nonlinear activations, demonstrating that fully connected architectures \cite{lee2018deep,g.2018gaussian,Hanin2023} and even some convolutional architectures \cite{novak2019bayesian, garriga-alonso2018deep} exhibit asymptotically Gaussian behaviour in the infinite-width limit. 
In this limit, Gaussian processes emerge naturally due to central limit effects on the network’s outputs, resulting in the so-called Neural Network Gaussian Process (NNGP).
Convergence rates to gaussian limit for  fully connected networks  have also been derived \cite{favaro2023quantitative, Trevisan2023}.
The simplification of random neural networks in the NNGP regime comes at a significant cost in terms of the model’s descriptive power. 
Specifically, for fully connected networks in the NNGP regime, the prior distribution yields independent output components and,  assuming   a Gaussian likelihood,   the   outputs are  independent also under  the posterior distribution.  In summary,  NNGP cannot capture {data}  dependent features. 
Moreover, while observed responses affect the posterior mean, they do not influence its covariance structure. 

This lack of learning capability is somewhat disappointing, especially given that modern deep architectures are known to perform complex feature learning beyond what is achievable in the infinite-width limit, see \cite{ChizatLazy, lewkowycz2021the, NEURIPS2020_1457c0d6}.
This limitation has oriented the research  in finding alternative limit regimes that allow non-Gaussian structures, potentially improving the network's capacity to learn data-driven features. 
The gaussian limit can be avoided with different approaches. For example using the mean field  scaling, see e.g.  \cite{doi:10.1073/pnas.1806579115,https://doi.org/10.1002/cpa.22074,doi:10.1137/18M1192184,NEURIPS2018_a1afc58c}, the  maximal update parametrizations \cite{yang2020feature},
heavy tailed initial weight distributions in a Bayesian setting, see e.g.  \cite{BordinoFavaroFortini2023,FavaroFortiniPeluchetti2023} or 
the so-called proportional limit (where both the number of training patterns $P$ and the number of neurons $N$ diverges at the same rate) investigated
in physics literature   \cite{pacelli2023statistical,aiudi2023,baglioni2024predictive}.

A recent line of research \cite{hanin2024} proposes to examine the ratio of depth $L$ (the number of layers) to width $N$ (the number of neurons per layer), referred to as the 'effective depth', in order to understand mechanisms that may drive neural networks toward a non-Gaussian asymptotic behaviour.  
In the proportional limit, where both depth and width diverge but their ratio $L/N$ converges to a positive constant $a>0$, 
the cumulant analysis in \cite{hanin2024} shows  that the limiting distribution cannot be Gaussian. In point of fact, for ReLU and linear activations, the limiting distribution has been proved to be   a mixture of Gaussians when considering a single input and a single output.

Despite their simplicity,  deep linear networks (i.e. networks with a linear activation) capture certain features of nonlinear deep networks
and attracted the interest of many  researchers, see e.g. \cite{Saxelinear, doi:10.1073/pnas.1820226116,SompolinskyLinear,doi:10.1073/pnas.2301345120,zavatone-veth2021exact,https://doi.org/10.1002/cpa.22200}.

Recently, \cite{BaRoetal} provide an  elementary representation for the  prior distribution over the outputs in a deep linear network, given in terms of a mixture of Gaussians. 
Building on this representaion,  in our paper we show that in the proportional limit $L/N \to a$ with $a>0$, linear networks with general input and output dimensions converge to a nontrivial Gaussian mixture, generalizing the above mentioned single input and single output results for this regime. We show that the mixing distribution can be  explicitly defined by multiple stochastic Brownian integrals (Proposition 
\ref{LNlimitwithKgendim}) and that  a similar representation holds also for the posterior distribution (Propositions \ref{Prop_limit_posterior} and 
\ref{prop2A}). 
Interestingly, the Gaussian mixture  limit allows the network to learn dependencies in the responses, a feature observed in finite networks but absent in the infinite-width limit. Moreover,  unlike in the NNGP regime, in the proportional limit the output correlations depend on the observed labels.  This makes the proportional limit  a closer approximation  to finite network behaviour  than the infinite-width setting.

\section{Setting of the problem}

\subsection{Bayesian fully-connected deep linear neural networks}

In a  {\it fully-connected linear neural network with $L$ hidden layers}, 
the pre-activations of each layer $h_{i_{\ell}}^{(\ell)}$ ($i_{\ell} = 1,\dots, N_{\ell}$; $\ell = 1, \dots, L+1$) are given recursively as a function of the pre-activations of the previous layer $h_{i_{\ell-1}}^{(\ell-1)}$ ($i_{\ell-1}= 1, \dots, N_{\ell-1}$).
For $\mathbf x=(x_1,\dots,x_{N_0})^\top$ 
\begin{equation}
    \label{main_recursion}
    \begin{split}
h_{i_1}^{(1)}(\mathbf x) &= \frac{1}{\sqrt {N_{0}}} \sum_{i_0=1}^{N_{0}} W^{(0)}_{i_1 i_0} x_{i_0} \, \\
h_{i_\ell}^{(\ell)}(\mathbf x) &= \frac{1}{\sqrt {N_{\ell-1}}} \sum_{i_{\ell-1}=1}^{N_{\ell-1}} W^{(\ell-1)}_{i_{\ell}i_{\ell-1}} h_{i_{\ell-1}}^{(\ell-1)}(\mathbf x) \,,
\qquad \ell=2,\dots,L+1
 \\
\end{split}
\end{equation}
where $W^{(\ell-1)}$ are  the weights and we assume that the so-called  biases of the $\ell$-th layer are zero.
Assuming  $N_{L+1}=D$,  the function implemented by the neural network   is  
 the output of the last layer $f_{L} (\mathbf x | \theta )=(h_{1}^{(L+1)}(\mathbf x),\dots,h_{D}^{(L+1)}(\mathbf x))^\top$.
  Here 
$\theta= \{W^{(\ell-1)}_{i_\ell,i_{\ell-1}}: \ell=1,\dots,L+1;  i_{\ell}=1,\dots, N_{\ell} \}$ 
represents the collection of all the trainable weights of the network.

Collecting   $P$ possible  inputs 
 $\bx_{\mu} \in \mathbb{R}^{N_0}$  ($\mu=1,\dots,P$)    
in a  $N_0 \times P$  matrix
$\bX=[\bx_{1},\dots,\bx_{P}]$, 
the function implemented by a deep linear neural network for these inputs  in matrix form 
has the  explicit expression
\begin{equation}\label{basic_rep_in_matrix_form}
   f_{L}(\bX| \theta) = [f_{L} (\mathbf x_1|\theta),\dots, f_{L} (\mathbf x_P|\theta)] =  \frac{W^{(L)}}{\sqrt{N_{L}}} \cdots
    \frac{W^{(0)}}{\sqrt{N_{0}}}
    \bX   
\end{equation}
where $W^{(\ell)}=[W_{i_{\ell+1},i_\ell}^{(\ell)} ]$ are $N_{\ell+1}\times N_{\ell}$ matrices.

In a Bayesian neural network, 
 a prior for the weights $\theta$ is specified, which 
translates in a prior for $f_{L} (\mathbf x_\mu|\theta)$. 
It is common to assume a gaussian prior for the weights, more precisely the $W^{(\ell)}_{ij}$ are independent and 
normally distributed with zero mean and layer dependent variance $\lambda_\ell^{-1}$, that is 
\begin{equation}\label{lawofW}
   W^{(\ell)}_{ij}  \stackrel{ind}{\sim}  \CN\Big(0,\lambda_\ell^{-1}\Big). 
 \end{equation}
 Here and in the rest of the paper $\CN(\mathbf{m},\mathbf{C})$ denotes the Gaussian distribution 
 with mean $\mathbf{m}$ and covariance matrix (variance) $\mathbf{C}$ (depending on the dimension $d>1$ or $d=1$). 

The prior over the outputs is the law of the random 
matrix   $f_{L}(\bX|\theta)$ or, equivalently, of the collection  of  random 
vectors  $\{f_{L} (\mathbf x_\mu|\theta); \mu=1,\dots,P\}$.

\subsection{Explicit prior  expression} 

In this section, we review  a result proved in \cite{BaRoetal}, which serves as the basis for our study. 
Denote by $\CS^+_{D}$ the set of $D \times D$  symmetric strictly positive definite matrices  and recall 
 that a random matrix  $Q$ taking values in  $\CS^+_D$ has  Wishart  distribution with $N>D$ 
 degrees of freedom and scale matrix $C$  if 
 $Q=\sum_{i=1}^N \bZ_i \bZ_i^\top$ where $\bZ_i$ are independent   Gaussian vectors in $\RE^D$
with zero mean and  covariance matrix $C$.
Equivalently, $Q$ has 
a Wishart distribution with $N$ degrees of freedom and scale matrix $C$ if  it has the following density
(with respect to  the Lebesgue measure on the cone of symmetric positive definite matrices)
\[
 Q \mapsto \frac{\det(Q)^{\frac{N-D-1}{2} }  e^{-\frac{1}{2}\tr(C^{-1}Q)}}{\det(C)^{N/2}2^{DN/2}(\pi)^{D(D-1)/4}\prod_{k=1}^D\Gamma\!\left(\frac{N - D +k}{2}\right)}.
 \]
See \cite{Eaton2007} and the references therein.  
If $D=1$, one has that $\CS^+_{D}=\RE^+$ and the Wishart distribution reduces to  a Gamma distribution of parameters $\alpha=\frac{N}{2}$, $\beta=\frac{1}{2C}$, 
that is  
\[
Q \mapsto \frac{1}{(2C)^{N/2} \Gamma(N/2)}   
Q^{\frac{N}{2}-1} e^{- \frac{1}{2C}Q} \quad \text{for $Q>0$}.
\]
It is possible to prove that  in a fully-connected linear network the prior distribution over $f(\bX|\theta)$ is a  mixture of Gaussians where the covariance matrix is
an explicit function of random Wishart matrices. 
From now on $Y_1 \stackrel{\CL}{=} Y_2$ means that the two random elements $Y_1$ and $Y_2$ have the same law, moreover
$Y_n \stackrel{\CL}{\to} Y$ means that the sequence of random elements $(Y_n)_n$ converges in law to $Y$ as $n \to +\infty$. 

\begin{proposition}[\cite{BaRoetal}]\label{prop1}   
Let  $ f_{L}(\bX| \theta)$ be 
the outputs  of a 
 fully-connected linear network  under the prior \eqref{lawofW}. 
If $\min(N_\ell:\ell=1,\dots,L) >D$ and $\lambda^*_L:=\lambda_0 \dots \lambda_L$, then 
\begin{equation}\label{main_rep}
 f_{L}(\bX|\theta) \stackrel{\CL}{=} V^{L} \cdots  V^{1}  \frac{\bZ }{\sqrt{N_0 \lambda^*_L}}   \bX
\end{equation}
where $V^{\ell}$ are any $D \times D$ independent random matrices
such that $Q^\ell= V^{\ell} (V^{\ell})^\top$
  has a Wishart distribution 
with $N_\ell$ degrees of freedom 
and scale matrix  $\frac{1}{N_\ell}\eI_D$ and 
 $\bZ$ is a $D \times N_0$ matrix of independent standard normal random variables. 
\end{proposition} 

Above and in the rest of the paper, $\eI_{M}$ is the identity matrix 
of dimension $M \times M$. 

It is worth noticing that  in Proposition \ref{prop1}, 
for $D \geq 2$,
one can choose any decomposition 
 $Q^\ell=V^\ell (V^\ell)^\top$. For convenience   in what follows  we choose the Cholesky decomposition 
where $V^\ell= \psi(Q^\ell)$ is a lower triangular matrix with positive diagonal entries. This $V^\ell$ is called Cholesky factor (or Cholesky square root).
 The Cholesky square root $\psi$ 
 is one to one and continuous
from $\CS_D^+$ to its image. 

When $Q^\ell$ has a Wishart distribution the law of its
Cholesky square root $V^\ell$ is the so-called
Bartlett distribution (see~\cite{kshirsagar1959}).
A  lower triangular random matrix   $V^\ell$ 
is said to have a 
Bartlett distribution
with $N_\ell$ degrees of freedom 
and scale matrix  $\frac{1}{N_\ell}\eI_D$  if
 \begin{equation}\label{def.Bartlett.matrix}
\begin{split}
&  V^{\ell}_{ii} \,\, \text{is such that}  \,\, (V^{\ell}_{ii})^2  \sim Gamma((N_\ell-i+1)/2,N_\ell/2 )  \quad  \text{ $i=1,\dots,D$}   \\
& V^{\ell}_{ki}       \sim  \mathcal{N}(0,N_\ell^{-1})  \quad       \text{for $D \geq k>i \geq 1$} \\
\end{split}
\end{equation}
and all the elements are independent. For the Gamma  distribution we use the shape-rate parametrization.

\vskip 0.5cm

{\bf Notation. } {\it The laws of the $V^{\ell}$'s  clearly depend on $N_\ell$, so that one should more correctly  write 
$V^{\ell}=V^{{\ell},N_\ell}$. For the sake of simplicity, we will sometimes  omit this dependence. 
The same observation  holds also for $Q^\ell$. 
When it will be  useful we shall write $Q^{\ell,N_\ell}$  in place of $Q^\ell$.
} 

\subsection{Infinite-width infinite-depth asymptotic regime} 

In the rest of the paper we shall  consider  the 
 proportional  infinite-width  infinite-depth
asymptotic regime in which we fix the input and the output dimension ($N_0$ and $N_{L+1}=D$) and 
we assume that   both the number of layers and the number of neurons diverge, more precisely:
\begin{equation}\label{H_asym}
\begin{split}
& N_1=N_2=\dots=N_L=N ,   \quad N\to +\infty,   \\
&L=L(N), \,\,\,  \lim_{N \to +\infty} L(N) = L_\infty \leq +\infty, \quad
 \lim_{N \to +\infty}   L(N)/N = a, \quad 0 \leq a <+\infty. \\
\end{split} 
\end{equation} 

In what follows, in order to stress the fact that  $ f_{L}(\bX|\theta)$  depends on $N$,  we shall  use the notation $ f_{L,N}(\bX|\theta)$.

Before considering the general case, we briefly review what happens in the so-called lazy-training infinite-width limit, that is, 
when $L(N)=L$ is a constant and $N \to +\infty$  (a special case 
of the previous setting with $a=0$).  
Since $Q^{\ell,N} \stackrel{\CL}{=}\frac{1}{N} \sum_{j=1}^N 
\bZ_{j,\ell} \bZ_{j,\ell}^\top$, where $\bZ_{j,\ell}$ are independent standard Gaussian vectors, the law of large numbers yields that $( Q^{\ell,N} ,\dots, Q^{L,N} )$ converges (in law) to $(\eI_D,\dots,\eI_D)$.
Using the fact that the map $(Q^{\ell} ,\dots,Q^{L} ) \mapsto V^L \cdots V^1$ is continuous, $V^L \cdots V^1$ also converges in probability to $\eI_D$. This gives 
the well-known NNGP infinite-width limit:

{\it If $L$ is fixed as $N \to +\infty$, 
\begin{equation*}
 f_{L,N}(\bX|\theta)  \stackrel{\CL}{\to} \frac{\bZ}{\sqrt{N_0 \lambda^*_L}} \bX,
\end{equation*}
where $\bZ$ is a $D \times N_0$ matrix of independent standard normal random variables. } 

Our main result is that this still holds if $a=0$, but as soon as $0<a<+\infty$, the asymptotic normality no longer holds, and the limit distribution becomes a non-trivial mixture of Gaussians.

By representation \eqref{main_rep} it is apparent that  the crucial point is to study the asymptotic distribution of the
following product of random matrices
\[
\bar V^{L,N}:=V^{L,N} \cdots V^{1,N}.
\]

\section{Main results}\label{Subsec:K=1}

To introduce the main results and ideas, we begin by considering the simple case where $D=1$.
 As recalled above, in this case  the $V^{\ell}$'s  in Proposition \ref{prop1} 
are independent random variables
 distributed as the square root of a $Gamma(N/2, N/2)$ 
 random variable. 
Hence, 
$\bar V^{L,N}$ is a scalar random variable 
and the prior distribution of the output is the same of the distribution of 
 $ {\bar V^{L,N}}  {\bZ }{(\sqrt{N_0 \lambda^*_L})^{-1}}   \bX $
 where $\bZ \sim \CN(\mathbf{0},\eI_{N_0})$. 
 
In this case, it is very easy to  describe the limiting distribution of   $\bar V^{L,N}$.
First of all one has 
\[
\bar V^{L,N}=\prod_{\ell=1}^L
V^{\ell,N}=\exp \left ( \frac{1}{2}  \sum_{\ell=1}^L\log(Q^{\ell,N} )\right ) 
\] 
where $(V^{\ell,N})^2=Q^{\ell,N}$  are independent  $Gamma(N/2,N/2)$ random variables. 

By direct computations, one can shows that for any real number $s$
\[
\lim_{N \to \infty,  L/N \to a} 
\E[e^{\frac{s}{2} \sum_{\ell=1}^L\log(Q^{\ell,N})}]=e^{\frac{a}{2} \frac{s^2}{2}-s\frac{a}{2}}.
\]
The right hand side is the moment generating function 
of a Normal distribution, which shows that 
$\frac{1}{2}\sum_{\ell=1}^L\log(Q^{\ell,N})$
converges in distribution to a normal random variable $Z_\infty$, with $Z_\infty \sim \CN(-\frac{a}{2},\frac{a}{2})$
if $a>0$ and $Z_\infty=0$ if $a=0$. Details are given in next Lemma \ref{conv.of.diag.elem}.
Using the continuous mapping theorem for convergence in distribution
 we obtain   the next statement which is the one-dimensional version of our main result. 

 {\it If  $D=1$ and 
  \eqref{H_asym} holds, 
then $\bar V^{L,N} $ converges in distribution
to a log-normal random variable 
$\bar V^{\infty}=\exp ( Z_\infty )$ where $Z_\infty \sim \CN(-a/2,a/2)$, with the convention $\mathcal{N}(0,0)=\delta_0$. 
If in addition,    $\lambda^*_L \to \lambda^*_\infty$ as $N \to +\infty$,  then 
\[
 f_{L,N}(\bX|\theta) \stackrel{\CL}{\to} {\bar V^{\infty}}  \frac{\bZ }{\sqrt{N_0 \lambda^*_\infty}}   \bX
\]
 where $\bZ$ is a vector of $N_0$ independent standard normal random variables, $\bar V^{\infty}$ and $\bZ$ being independent. 
}

The previous result shows  that in the proportional  infinite-width  infinite-depth limit  \eqref{H_asym},  provided that $a>0$, the asymptotic  
 prior is a  non-degenerate 
mixture of Gaussians.
This  is essentially already known, and it is contained in equation (76) of  Section B.2.1 in \cite{hanin2024}, 
although  \cite{hanin2024}  deals only with the case of one input, i.e.   $P=1$ and  $\bX=\bx_1$.

Our main result is that  this can be extended  
also to the more general case  $D > 1$ as well to the posterior distribution of the outputs, as we shall see in the next subsections.

\subsection{Prior asymptotics}

If    $a=0$ in   \eqref{H_asym}, the limit prior   of the network 
is the same as in the NNGP regime, as shown in the next proposition. 

\begin{proposition}[the case $a=0$ and $D \geq 1$]\label{Prop:lazy}
  If  \eqref{H_asym} holds with $a=0$, 
that is  $L/N \to 0$,   the sequence of  random matrices  $\bar V^{L,N}$  
 converges in probability 
 to the identity matrix.  
If in addition  $\lambda^*_L \to \lambda^*_\infty$ as $N \to +\infty$,  then 
\begin{equation*}
 f_{L,N}(\bX|\theta)  \stackrel{\CL}{\to} \frac{\bZ}{\sqrt{N_0 \lambda^*}} \bX,
\end{equation*}
where $\bZ$ is a $D \times N_0$ matrix of independent standard normal random variables. 
\end{proposition}

In order to describe the limit for $a>0$, we introduce some additional notation. 
Let $(W^{(i,j)}_t)_{ t \in [0,1]}$  and $(W_t^{(k)})_{t \in [0,1]}$ with $1 \leq j <i \leq D$ and $k=1,\dots,D$   be independent standard  Brownian    motions
and set 
\begin{equation}\label{def.Zkt}
Z^{(k)}_t:=\sqrt{\frac{a}{2}}W^{(k)}_t-\frac{k}{2}a t \qquad  \text{for $k=1,\dots,D$}. 
\end{equation}
For $1 \leq i < k \leq D$ and $h=1,\dots,k-i$, 
set  $\CR_{k,i}^h=\{ \br=(r_0,\dots,r_h):r_0:=i<r_1<\cdots<r_{h-1}<r_h:=k\}$
and for $\br \in \CR_{k,i}^h$ 
define the multiple stochastic integral
\begin{equation}\label{multiplestochasticintegral}
\begin{split}
H({\br}) & :=e^{\frac{a}{2}\ln(a)+ Z^{(r_h)}_1} \int_0^1   e^{Z^{(r_0)}_{t_{1}}-Z^{(r_{1})}_{t_{1}}  } \int_{t_1}^1 e^{Z^{(r_1)}_{t_{2}}-Z^{(r_{2})}_{t_{2}}  } 
\\
&  \cdots 
\int_{t_{h-1}}^1  e^{Z^{(r_{h-1})}_{t_h}-Z^{(r_h)}_{t_h}  }  dW^{(r_{h},r_{h-1})}_{t_{h}}
\cdots
     dW^{(r_2,r_1)}_{t_2}     dW^{(r_1,r_0)}_{t_1} ,  \\
     \end{split}
\end{equation} 
(with the convention that $t_0=0$ if $h=1$ and the multiple integral reduces to a single integral). The previous stochastic integrals are
well-defined, see next  Section \ref{Sec:filtrations} and Lemma \ref{lemma_wellposedness} for details. 
Finally, introduce the random lower triangular  matrix $\bar V^{\infty}$ by
\begin{equation}\label{def:Vinf}
\begin{split}
& \bar V^{\infty}_{k,i} : = \sum_{h=1}^{k-i} \sum_{ \br \in \CR_{k,i}^h } H({\br})  \quad 1 \leq i < k \leq D \\
& \bar V^{\infty}_{k,k}:= e^{ Z^{(k)}_1} \quad k=1,\dots,D. \\
\end{split}
\end{equation}

\begin{proposition}[the case $a>0$ and $D \geq 1$]\label{LNlimitwithKgendim}
Let $D\geq 1$ and 
assume  \eqref{H_asym} with $a>0$. 
  Then as $N \to +\infty$ 
  \[
   \bar V^{L,N}
\stackrel{\mathcal{L}}{\to} 
\bar V^{\infty}.
  \]
If in addition   $\lambda^*_L \to \lambda^*_\infty$ as $N \to +\infty$,  then  
\begin{equation*}
 f_{L,N}(\bX|\theta)  \stackrel{\CL}{\to} \bar V^{\infty} \frac{\bZ}{\sqrt{N_0 \lambda^*}} \bX,
\end{equation*}
where $\bZ$ is a $D \times N_0$ matrix of independent standard normal random variables, 
$\bar V^{\infty}$ and $\bZ$ being independent. 
\end{proposition} 

\subsection{Posterior  asymptotics}\label{Sec:BayesianStting}

In a  supervised learning problem one has  a training set $\{\mathbf x_\mu ,\mathbf y_\mu\}_{\mu=1}^P$, where each $\mathbf x_\mu \in \mathbb R^{N_0 }$ 
has  the corresponding labels (response) $\mathbf y_\mu \in \RE^D$. 
Setting   $\bS_{\mu}:= f_{L} (\mathbf x_\mu|\theta)$ for  $\mu=1,\dots,P$, 
the prior  over the outputs   is  the law under \eqref{lawofW}  of  
$\{ \bS_{\mu}; \mu=1,\dots,P\}$. 
In order to perform Bayesian learning for the network parameters, or directly for the network outputs, 
one requires a {\it  likelihood for the labels given the inputs and the outputs}, which we will denote by
 $\CL(\by_{1},\dots, \by_P|\bs_{1},\dots,\bs_{P})$. In probabilistic terms, the function
\[
(\by_{1},\dots, \by_P) \mapsto \CL(\by_{1},\dots, \by_P|\bs_{1},\dots,\bs_{P})
\] represents the conditional density of the response given the inputs  $\bS_1=\bs_{1},\dots,\bS_P=\bs_{P}$.
For example, in analogy to a network trained with a quadratic loss function, one can consider a Gaussian likelihood
\begin{equation}\label{gaussian_likelihood}
\CL(\by_{1},\dots, \by_P|\bs_{1},\dots,\bs_P)  \propto e^{-\frac{\beta}{2}\sum_{\mu=1}^P \|\bs_{\mu}-\by_{\mu}\|^2}, 
\end{equation}
with $\beta>0$.  Note that this corresponds to assume the error model:
\[
\by_\mu=\bS_\mu+\bm{\varepsilon}_\mu
=f_L(\bx_\mu|\theta)+\bm{\varepsilon}_\mu \quad \bm{\varepsilon}_\mu \stackrel{iid}{\sim} \CN(\mathbf{0},\beta^{-1}\eI_D).
\]

The core of Bayesian learning is captured by the posterior distribution  of $f_{L}(\bX|\theta)$, i.e. 
the conditional distribution of  $f_{L}(\bX|\theta)$ 
given $\bY=[\by_{1},\dots,\by_P]$. 
To cover a slightly more general situation, it is convenient to introduce  an additional input  $\bx_0$ and the corresponding output 
$f_{L} (\mathbf x_0|\theta)$, which represents the output of the test set or  the output of  a new data input. Hence, 
while $\bY=[\by_1,\dots,\by_P]$ are available, $\by_0$ is not observed. 
If one sets  $\tilde \bX=[\bx_0,\bx_1,\dots,\bx_P]=[\bx_0,\bX]$, then 
the collection of all the outputs is the $D (P+1)$ matrix 
$f_{L,N}(\tilde {\mathbf X}|\theta)$ defined in 
\eqref{basic_rep_in_matrix_form} with $\tilde \bX$ in place of $\bX$.

In what follows, it is useful to transform the random matrix $f_{L,N}( \tilde {\mathbf X}|\theta)$ in a vector, 
defining   $\bS_{N,0:P}=\vvec[f_{L,N}( \tilde {\mathbf X}|\theta)]$, where $\vvec[A]$  is  the operation of stacking the columns of matrix 
$A$ into a column  vector. From now on, $P_{N,\mathrm{prior}}(\cdot| \tilde \bX)$ denotes the  (prior) 
distribution of  $\bS_{N,0:P}$, that is 
by Proposition \ref{prop1} 
  \[
P_{N,\mathrm{prior}}(\cdot| \tilde \bX)=  \textbf{Law}((N_0 \lambda^*_L)^{-\frac{1}{2}}\vvec[\bar V^{L,N} \bZ   \tilde  \bX]).
  \]
Moreover,  Proposition \ref{LNlimitwithKgendim} (applied to the input $\tilde \bX$ in place of $\bX$) states that, 
if   \eqref{H_asym} holds and  $\lambda^*_L \to \lambda^*_\infty$, then 
$P_{N,\mathrm{prior}}(\cdot | \tilde \bX)$  converges weakly to 
\[
P_{\infty,\mathrm{prior}}(\cdot | \tilde \bX):=\textbf{Law}\Big ( (N_0 \lambda^*_\infty)^{-\frac{1}{2}} \vvec[V^{\infty} \bZ \tilde \bX]\Big ).
\]

Assuming that the likelihood  is bounded and continuous, 
this convergence easily translates from prior to posterior. 
This follows noticing that the joint posterior distribution 
 of  
$\bS_{N,0:P}$ is by Bayes theorem 
 \begin{equation}\label{posterior_N}
P_{N,\mathrm{post}}(d\bs_{0:P}| \bY, \tilde \bX) 
\propto 
\CL(\by_1,\dots,\by_P|\bs_{1:P} )    P_{N,\mathrm{prior}}(d\bs_{0:P}| \tilde \bX) 
 \end{equation}
where 
$\bs_{0:P}=\vvec[{\bs_0},\bs_{1},\dots,\bs_P]$, $\bs_{1:P}=\vvec[\bs_{1},\dots,\bs_P]$ for 
$\bs_i \in \RE^{D}$, $i=0,\dots,P$. We are assuming that, conditionally on $ \bS_1,\dots, \bS_N$, 
the distribution of the labels is independent from $ \bS_0$. 

Proposition \ref{LNlimitwithKgendim} gives  immediately the next result.

\begin{proposition}\label{Prop_limit_posterior}
If $\bs_{1:P} \mapsto \CL(\by_1,\dots,\by_P|\bs_{1:P})$ is a bounded and continuous function,   \eqref{H_asym} holds and  $\lambda^*_L \to \lambda^*_\infty$, then 
$P_{N,\mathrm{post}}( \cdot | \bY, \tilde \bX)$ converges weakly to 
 \begin{equation}\label{posterior_infty} 
P_{\infty,\mathrm{post}}(d\bs_{0:P}| \bY, \tilde \bX) 
=\frac{ \CL(\by_1,\dots,\by_P| \bs_{1:P} ) P_{\infty,\mathrm{prior}}(d\bs_{0:P}| \tilde \bX) 
}{ \int_{\RE^{D  (P+1)} } \CL(\by_1,\dots,\by_P|\bs ) P_{\infty,\mathrm{prior}}(d\bs |\tilde \bX)  }. 
\end{equation}
\end{proposition}

Marginalizing one obtains the posterior distribution of $f_{L} (\mathbf x_0|\theta)$  and the corresponding asymptotic result. 
Indeed, the  posterior predictive is the law  of $ \bS_0=f_{L} (\mathbf x_0|\theta)$  given $\bY$, 
easily  obtained   by
 \begin{equation}\label{predictive_0}
 \begin{split}
P_{N,\mathrm{pred}}(d\bs_0|\bY,\tilde \bX) & =\int_{\RE^{D P}}  P_{N,\mathrm{post}}(d\bs_0 d\bs_{1} \cdots d\bs_P| \bY, \tilde \bX) 
\\
&  \propto
\int_{\RE^{D P}} \CL( \by_{1},\dots,\by_P|\bs_{1},\dots,\bs_P)
P_{N,\mathrm{prior}}(d\bs_0 d\bs_{1} \cdots d\bs_P|\tilde \bX) \\
\end{split}
\end{equation}
where the integrals are taken only with respect to the variables $\bs_{1}, \dots, \bs_P$.

\subsection{Explicit expression of the asymptotic posterior under a gaussian likelihood}

If  the Gaussian likelihood \eqref{gaussian_likelihood} is assumed,
both  the  posterior   \eqref{posterior_N} and the asymptotic  posterior   \eqref{posterior_infty}
 turn out to be  mixtures of normal distributions. 
 
 For the sake of simplicity, in this section we assume that  $\lambda^*_L=\lambda^*_\infty=1$.  
As for the prior is concerned, standard manipulation of matrix normal random variables (see \ref{AA} and \ref{BB} in Appendix \ref{appendixA}), shows that 
the conditional distribution of  $N_0^{-\frac{1}{2}}\vvec[\bar V^{L,N} \bZ   \tilde  \bX]$ given 
 $\bar V^{L,N}$ is a normal distribution with mean $\mathbf{0}$ and covariance $N_0 ^{-1}   \tilde \bX^\top \tilde \bX \otimes \bar Q^{L,N}$
 where
\[
\bar Q^{L,N}:= \bar V^{L,N}  (\bar V^{L,N} )^\top
\]
and
 $A \otimes B$ is the 
Kroneker product of matrices $A$ and $B$.  Since $\bS_{N,0:P}\stackrel{\CL}{=} N_0^{-\frac{1}{2}}\vvec[\bar V^{L,N} \bZ   \tilde  \bX]$, 
 one has 
  \[
 P_{N,\mathrm{prior}}(d \bs_{0:P}| \tilde \bX)=   \int_{\CS^+_{D}}    
    \CN(d\bs_{0:P}| \mathbf{0}, N_0^{-1} \tilde\bX^\top \tilde\bX \otimes Q)
\mathcal{Q}_{L,N}(dQ)
 \]
 where  $\mathcal{Q}_{L,N}$ is the law of  $ \bar Q^{L,N}$  and   $\CN(\cdot |\mathbf{m},\mathbf{C})$
 denotes the distribution of a multivariate normal random vector  with mean $\mathbf{m}$ and covariance matrix $\mathbf{C}$. 
The same result holds for the limit 
$N_0^{-\frac{1}{2}}\vvec[\bar V^{\infty} \bZ   \tilde  \bX]$ 
with  
\[
\bar Q^{\infty}:=\bar V^\infty (\bar V^\infty)^\top
\]
 in place of $\bar Q^{L,N}$ and  $\mathcal{Q}_{\infty}(\cdot)$, which  is  the law of  $\bar Q^{\infty}$, in place of $\mathcal{Q}_{L,N}$. 

Assuming  the gaussian likelihood \eqref{gaussian_likelihood},  
 if   $\by_{1:P}=\vvec[\bY]$,
 the posterior distribution $P_{N,\mathrm{post}}$ is 
\begin{equation}\label{Post_N-bayes1}
P_{N,\mathrm{post}}(d\bs_{0:P}| \bY, \tilde \bX) 
  \propto 
  e^{-\frac{\beta}{2} \|\bs_{1:P}-\by_{1:P}\|^2}
    \int_{\CS^+_{D}}    
    \CN(d\bs_{0:P}| \mathbf{0}, N_0^{-1} \tilde\bX^\top \tilde\bX \otimes Q)
\mathcal{Q}_{L,N}(dQ). 
\end{equation}
Simple (but long)  computations show   that also $P_{N,\mathrm{post}}$  is   mixture of  multivariate normal distributions.  Exactly the same considerations hold for $P_{\infty,\mathrm{post}}(\cdot | \bY, \tilde \bX) $. 

 In order to fully describe these mixtures, we need some additional notation. 
Given ${\tilde{\bX}}=[\bx_0,\bX]$,  define the functions $\Sigma:\CS^+_{D} \to \CS^+_{D (P+1)}$
and $\Sigma^{*}:\CS^+_{D} \to \CS^+_{D (P+1)}$ 
 by 
\begin{equation}\label{def_sigmaQ}
\Sigma(Q|\tilde{\bX})=\begin{pmatrix}
\Sigma_{00}(Q|\tilde{\bX}) &  \Sigma_{01}(Q|\tilde{\bX}) \\
\Sigma_{01}^\top(Q|\tilde{\bX}) &  \Sigma_{11}(Q|\tilde{\bX}) \\
\end{pmatrix}
=
\frac{1}{N_0}
\begin{pmatrix}
{\bx_0^\top \bx_0 } \otimes Q  &   {\bx_0^\top \bX} \otimes Q \\
 { \bX^\top \bx_0} \otimes Q^\top & {\bX^\top \bX} \otimes Q 
  \\
\end{pmatrix}
\end{equation}
and
 \[
 \Sigma^{*}(Q |\tilde{\bX}):=\begin{pmatrix}
\Sigma^{*}_{00}(Q|\tilde{\bX}) &  \Sigma^{*}_{01}(Q|\tilde{\bX}) \\
\Sigma^{* \top }_{01}(Q|\tilde{\bX})&    \Sigma_{11}^{*}(Q|\tilde{\bX}) \\
\end{pmatrix}
\]
where 
\[
\begin{split}
 \Sigma^{*}_{11}(Q| \tilde \bX): & =\Sigma_{11} (Q| \tilde \bX) (\eI_{DP} +\beta \Sigma_{11}(Q| \tilde \bX))^{-1} \\
 \Sigma^{*}_{00}(Q| \tilde \bX): & = \Sigma_{00}(Q| \tilde \bX)-
 \Sigma_{01}(Q| \tilde \bX) \Sigma^{-}_{11}(Q| \tilde \bX)   \Big ( 
\eI_{DP} - \Sigma^{*}_{11}(Q| \tilde \bX)   \Sigma^{-}_{11}(Q| \tilde \bX)  
\Big )  \Sigma_{01}^\top(Q| \tilde \bX) 
 \\
 \Sigma^{*}_{01}(Q| \tilde \bX):& =
\Sigma_{01} (Q| \tilde \bX) \Sigma^{-}_{11} (Q| \tilde \bX)  \Sigma^{*}_{11}(Q| \tilde \bX)
\end{split} 
\]
and $\Sigma_{11}^-$ is the Moore-Penrose inverse of $\Sigma_{11}$. 
Given $\by_{1:P}$,  introduce  also  the functions 
 $\mathbf{m}^*:\CS^+_{D}  \to \RE^{D (P+1)}$ 
and $\Phi:\CS^+_{D} \to \RE$ 
by 
\[
\begin{split}
& \mathbf{m}^*(Q|\tilde{\bX},\by_{1:P}) := 
\begin{pmatrix}
\beta  \Sigma_{01}(Q| \tilde \bX )  \Sigma^{-}_{11} (Q| \tilde \bX) 
  \Sigma^{*}_{11}(Q| \tilde \bX)   \by_{1:p}\\
  \beta \Sigma^{*}_{11}(Q| \tilde \bX)   \by_{1:p} \\
\end{pmatrix}\\
 & \Psi(Q|\tilde{ \mathbf{X}},\by_{1:P}):=
\beta \by_{1:P}^\top (  \eI_{DP}+ \beta \Sigma_{11}(Q| \tilde \bX))^{-1} \by_{1:P}
 +\log(\det(\eI_{DP} +\beta \Sigma_{11}(Q|\tilde{ \mathbf{X}})) . \\
 \end{split}
\]

Finally, 
define the  probability measure   
\begin{equation}\label{def_Q_posterior}
 \mathcal{Q}_{L,N} (dQ |\tilde \bX, \by_{1:P})   :=
 \frac{e^{-\frac{1}{2} \Psi(Q|\tilde{\bX},\by_{1:P}) }\mathcal{Q}_{L,N} (dQ)  }{\int_{\CS^+_{D}}e^{-\frac{1}{2} \Psi(Q|\tilde{\bX},\by_{1:P}) } \mathcal{Q}_{L,N} (dQ) }  
 \end{equation}
 and  $\mathcal{Q}_{\infty} (dQ |\tilde \bX, \by_{1:P})$ be defined as above with $ \mathcal{Q}_{\infty}$ in place of $ \mathcal{Q}_{L,N}$. 
We are now in a position to write the explicit  expression of the posterior distribution. 

\begin{proposition}\label{prop2A}
Assume \eqref{gaussian_likelihood},    $\lambda^*_L=1$  and hence 
$\lambda^*_\infty=1$, 
then  
\begin{equation*}\label{posterior_finteL}
P_{N,\mathrm{post}}(d\bs_{0:P}| \bY, \tilde \bX) =
\int_{\CS^+_{D}} 
\CN \Big( d \bs_{0:P} \Big| \mathbf{m}^* (Q|\tilde{\bX},\by_{1:P}) ,\Sigma^{*}(Q| \tilde{\bX})\Big)
 \mathcal{Q}_{L,N}(dQ| \tilde{\bX}, \by_{1:P}) 
\end{equation*}
and
\begin{equation*}\label{posterior_infinteL}
P_{\infty,\mathrm{post}}(d \bs_{0:P}| \bY, \tilde \bX) =
\int_{\CS^+_{D}} 
\CN \Big(d  \bs_{0:P} \Big| \mathbf{m}^* (Q|\tilde{\bX},\by_{1:P}) ,\Sigma^{*}(Q| \tilde{\bX})\Big)
 \mathcal{Q}_{\infty}(dQ| \tilde{\bX}, \by_{1:P}). 
\end{equation*}
Morevoer,  $\mathcal{Q}_{L,N}(\cdot| \tilde{\bX}, \by_{1:P})$ and   $\mathcal{Q}_{\infty}(\cdot| \tilde{\bX}, \by_{1:P})$
are the posterior distributions of $ \bar Q^{L,N}$ and $\bar Q^{\infty}$ given $\by_{1:p}$. 
\end{proposition}

For a detailed derivation of the previous expression one can follows the same line of the proof of Proposition 12 in   \cite{BaRoetal}.
For the sake of completeness we report the full derivation in Appendix  \ref{AppendixMixture}.

From the previous proposition it follows that
the weak limit of $P_{N,\mathrm{pred}}(d\bs^0|\bY,\tilde \bX)$ is 
\[
P_{\infty,\mathrm{pred}}(d\bs_{0}| \bY, \tilde \bX) =
\int_{\CS^+_{D}} 
\CN \Big( d \bs_{0} \Big| \mathbf{m}_0^* (Q|\tilde{\bX},\by_{1:P}) ,\Sigma^{*}_{00}(Q| \tilde{\bX})\Big)
 \mathcal{Q}_{\infty}(dQ| \tilde{\bX}, \by_{1:P}) 
\]
where $\mathbf{m}_0^* (Q|\tilde{\bX},\by_{1:P}) :=\beta  \Sigma_{01}(Q| \tilde \bX )  \Sigma^{-}_{11}(Q| \tilde \bX)   \Sigma_{11}^*(Q| \tilde \bX)   \by_{1:p}$.

\begin{remark}
 When  $\det(\tilde \bX^\top \tilde \bX)>0$ both  $P_{N,\mathrm{prior}}( \cdot |\tilde \bX)$ and $P_{\infty,\mathrm{prior}}( \cdot | \tilde \bX)$
  have a density with respect to the Lebesgue measure.  
 To see this,  observe that  $\det(V^{\ell,N})>0$ almost surely for any $\ell$, which yields that $\det(\bar V^{L,N})>0$ and hence $\det(\bar Q^{L,N})>0$.
Similarly, by \eqref{def:Vinf},  one has $\det(\bar V^{\infty})>0$ with probability one, and hence also  $\det(\bar Q^{\infty})>0$. In summary,  
 both $\bar Q^{L,N}$ and  $\bar Q^{\infty}$ are strictly positive definite with probability one. 
Hence, if  $\det(\tilde \bX^\top \tilde \bX)>0$, using 
 $\det(\tilde\bX^\top \tilde\bX \otimes  \bar Q^{L,N} )=\det(\tilde\bX^\top \tilde\bX )^D \det(\bar Q^{L,N})^P$, 
 one gets that also  $\tilde\bX^\top \tilde\bX \otimes  \bar Q^{L,N}$ is   strictly positive definite with probability one.
  Analogously, one gets that 
 $\tilde\bX^\top \tilde\bX \otimes  \bar Q^{\infty}$ is   strictly positive definite with probability one. 
 In addition, as a consequence of what mentioned above, 
 from \eqref{posterior_N} and \eqref{posterior_infty}
one immediately gets that   also 
$P_{N,\mathrm{post}}(\cdot| \bY, \tilde \bX)$  and $P_{\infty,\mathrm{post}}(\cdot| \bY, \tilde \bX)$
are absolutely continuous. Moreover, 
 $\Sigma_{11}^-=\Sigma_{11}^{-1}$ and some simplifications in the expression of $\Sigma^*$ and $\mathbf{m}$
 occur.  In particular 
 \[
 \Sigma_{01}^*(Q| \tilde \bX)=\Sigma_{01}(Q| \tilde \bX)(\eI_{DP} +\beta \Sigma_{11}(Q| \tilde \bX))^{-1},
 \] 
 \[
 \Sigma_{00}^*(Q| \tilde \bX)= \Sigma_{00}(Q| \tilde \bX)- \Sigma_{01} (Q| \tilde \bX) (\eI_{DP} +\beta \Sigma_{11}(Q| \tilde \bX))^{-1}  \Sigma_{01}(Q| \tilde \bX)^\top
 \]
 and
 \[
 \mathbf{m}_0^* (Q|\tilde{\bX},\by_{1:P}) =\beta \Sigma_{01}(Q|\tilde{\bX}) (\eI_{DP} +\beta \Sigma_{11}(Q| \tilde \bX))^{-1}  \by_{1:p}. 
\]
\end{remark}

A few comments are in order. 
If $a=0$, for example in the NNGP regime, the limit of the prior over the outputs  is 
\[
P_{\infty,\mathrm{prior}}(\cdot | \tilde \bX)=\textbf{Law}\Big ( N_0^{-\frac{1}{2}} \vvec[ \bZ \tilde \bX]\Big ),
\]
which is the same of saying that $\bar V^\infty=\bar Q^{\infty}=\eI_D$  and hence $\mathcal{Q}_{\infty}=\delta_{\eI_{D}}$.   
This yields that the posterior $P_{\infty,\mathrm{post}}(\cdot| \bY, \tilde \bX)$  is a  multivariate normal distribution and not a mixture of normal distributions. 
As a consequence, %
when $a=0$, the covariance is given by  $\frac{\tilde \bX^\top \tilde \bX}{N_0  } \otimes \eI_{D}$, which is completely independent of the data labels 
$\by_{1:P}$'s. Furthermore,  the output components are independent.   
In contrast, when  $a>0$, the  covariance of the outputs is 
 \[
 \Cov\Big(N_0^{-\frac{1}{2}}\vvec[\bar V^{L,N} \bZ   \tilde  \bX]\Big |\by_{1:p}\Big)=
 \int_{\CS^+_{D}} \Big ( \Sigma^{*}(Q| \tilde{\bX}) + \mathbf{m}^* (Q|\tilde{\bX},\by_{1:P})  \mathbf{m}^{* \top} (Q|\tilde{\bX},\by_{1:P})\Big )
 \mathcal{Q}_{\infty}(dQ| \tilde{\bX}, \by_{1:P}) 
 \]
 where the posterior $\mathcal{Q_\infty}(dQ |  \tilde{\bX}, \by_{1:P})$ explicitly  depends on  $\by_{1:P}$ through $\Psi(Q|\tilde{\bX},\by_{1:P})$, see
 \eqref{def_Q_posterior}. 
  The dependence of the covariance on the labels shows a learning process which is absent in the NNGP  infinite-width  limit. 
  Moreover, 
in the proportional limit, the output are not independent.  
In this sense, the proportional limit  is closer to 
 finite network behaviour than the infinite-width setting.

\section{Proofs. }

\subsection{Preliminary computations and proof of Proposition \ref{Prop:lazy}}
For $k \geq i$ the  $(k,i)$-element of the
(lower triangular) 
product matrix $\bar V^{L,N}=V^L \cdots V^1$
 is 
 \begin{equation}\label{barVkjA}
\bar V^{L,N}_{k,i} = 
\sum_{i \leq  j_1 \leq  \dots \leq j_{L-1} \leq k}
 V^1_{j_{1},i} V^2_{j_{2},j_1} \cdots
 V^{L}_{k, j_{L-1}}  
\end{equation} 
and, in particular,   the diagonal elements are 
 \begin{equation}\label{barVkjB}
\bar V^{L,N}_{r,r} = 
 V^{1}_{r,r} V^{2}_{r,r}   \dots V^{L}_{r,r}  
 =e^{\sum_{\ell=1}^L   \log(V^{\ell}_{r,r}) }. 
\end{equation}

We first consider the diagonal elements. 

\begin{lemma}\label{conv.of.diag.elem}
 If \eqref{H_asym} holds, as $N \to +\infty$
\[
(\bar V^{L,N}_{1,1},\dots,\bar V^{L,N}_{D,D})
 \stackrel{law}{\to}  (e^{Z_1},\dots,e^{Z_D})
\]
with $Z_1,\dots,Z_D$ independent  and $Z_r \sim \mathcal{N}(-a r/2,a/2)$ ($r=1,\dots,D$), with the convention $\mathcal{N}(0,0)=\delta_0$. 
\end{lemma}

\begin{proof}
If $X=\log(G)$ where $G \sim Gamma(\alpha,\beta)$, 
it is not difficult so see that   
\begin{equation}\label{MG-logGamma}
\E[  e^{s X }]  =\frac{\Gamma(\alpha+s)}{\Gamma(\alpha)} \beta^{-s}   \quad s> -\alpha  .
\end{equation}
Moreover, 
\begin{equation}\label{gammaexpansion}
 \Gamma(x+\alpha)/\Gamma(x)=
x^\alpha(1+\alpha(\alpha-1)/2x+O(x^{-2})) \quad \text{for $x \to +\infty$},
\end{equation}
see, e.g.,  \cite{Tricomi1951}. 
Recalling \eqref{def.Bartlett.matrix},  $(V^{\ell,N}_{r,r})^2 \sim  Gamma((N-r+1)/2,N/2 )$, 
so that by \eqref{MG-logGamma} and \eqref{gammaexpansion} one has 
\[
\begin{split}
\E[e^{s \log(\bar V^{L,N}_{r,r})}]& =
\E[e^{s \sum_{\ell=1}^{L}\frac{1}{2} \log\big ( (V^{\ell,N}_{r,r})^2 \big) } ]=
\left (  
\frac{\Gamma(\frac{N-r+1}{2}+\frac{s}{2})}{\Gamma(\frac{N-r+1}{2})} \Big (\frac{2}{N} \Big)^{\frac{s}{2} }   
\right)^{L} \\
& 
=\Big(1- \frac{r-1}{N} \Big)^L\Big(1+\frac{s(s-2)}{4(N-r+1)}+O(N^-2)\Big)^L.
\\
\end{split}
\]
Hence
\[
\lim_{N \to +\infty} \E[e^{s \log(\bar V^{L,N}_{r,r})}] = 
\exp \Big \{ \frac{a}{2}\frac{s^2}{2}-s\frac{ra}{2} \Big  \}=\E[e^{s Z_r}] .
\]
Since $ \bar V^{L,N}_{r,r}=\exp( \log(\bar V^{L,N}_{r,r}))$,
the thesis follows by  the continuous mapping theorem 
and the independence of the $ \bar V^{L,N}_{r,r}$'s.
\end{proof}

\begin{lemma}\label{Lemma1}
For $N>D$ and 
for $1 \leq i<k \leq D$ one has 
$\E[\bar V^{L,N}_{k,i}]=0$
and, for  $i<k \leq D$,
\[
\Var(\bar V^{L,N}_{k,i})= \E[(\bar V^{L,N}_{k,i})^2] \leq \sum_{m=1}^{k-i} \frac{1}{N^m}
{L \choose m} {k-i-1 \choose m-1}.
\]
Hence,  if \eqref{H_asym} holds,
then $\sup_{k=1,\dots,D, i <k,  N > D}\E[(\bar V^{L,N}_{k,i})^2]  <+\infty$
and, if \eqref{H_asym} holds
with $a=0$,
then $\lim_{N \to +\infty} \E[(\bar V^{L,N}_{k,i})^2] = 0$. 
\end{lemma}

\begin{proof}
By independence 
$\E[\bar V^{L,N}_{k,i} ]=
\sum_{k \geq j_{L-1} \dots \geq j_1 \geq i}  \E[V^L_{k , j_{L-1}} ]  \cdots \E[V^1_{j_1,i}]=0$
since $\E[V^\ell_{j,j'}]=0$ for $j \not = j'$, and  in any 
sequence $j_{L}:=k \geq j_{L-1} \dots \geq j_1 \geq i=:j_0$  there is at least 
 one $(j_{\ell+1},j_\ell)$ with $j_{\ell+1}\not= j_\ell$, since $k>i$. 
Now write $\mathbf{j}^{(r)}=(j_1^{(r)},j_{2}^{(r)},\dots,j_{L-1}^{(r)})$ 
with $k \geq j_{L-1}^{(r)}\dots \geq j_1^{(r)} \geq i$ for $r=1,2$ and 
\[
\begin{split}
\E[(\bar V^{L,N}_{k,i} )^2]&  =
\sum_{\bold{j}^{(1)},\bold{j}^{(2)}} 
\E[ V^L_{k, j_{L-1}^{(1)}} V^L_{k , j_{L-1}^{(2)}}    
  \dots V^1_{ j_1^{(1)},i} V^1_{j_1^{(2)},i}] \\
& = \sum_{\bold{j}^{(1)} \not =\bold{j}^{(2)}} 
\E[ V^L_{k,  j_{L-1}^{(1)}} V^L_{k , j_{L-1}^{(2)}}] 
\E[ V^{L-1}_{j_{L-1}^{(1)},  j_{L-2}^{(1)}} V^{L-1}_{j_{L-1}^{(2)} ,j_{L-2}^{(2)}}]   \cdots \E[ V^1_{ j_1^{(1)},i}
   V^1_{ j_1^{(2)},i} ] \\
&  +\sum_{\bold{j}^{(1)}  } 
\E[ (V^L_{k , j_{L-1}^{(1)}})^2 ]  \cdots \E[ (V^1_{j_1^{(1)},i})^2].
  \\
\end{split}
\]
Using the fact that the $V^\ell_{i_1,i_2}$'s are independent and that 
  $\E[V^\ell_{i_1,i_2}]=0$ if $i_1 \not =i_2$,  one can check that 
  if $\bold{j}^{(1)} \not =\bold{j}^{(2)}$ then 
  \[
\E[ V^L_{k,  j_{L-1}^{(1)}} V^L_{k , j_{L-1}^{(2)}} ]  \cdots  \E[  V^1_{j_1^{(1)},i}
   V^1_{j_1^{(2)},i} ]=0.
  \]
Now,  
 $\E[(V^{\ell}_{r,r})^2]=(N-r+1)/N\leq 1$ and 
$\E[(V^\ell_{j,j'})^2]=1/N$ if $j>j'$. Hence,  
given $\bold{j}^{(1)}$, if  $m$ is the number of pairs $(j_r^{(1)},j_{r+1}^{(1)})$ 
for which $j_r^{(1)} \not =j_{r+1}^{(1)}$, one has that 
\[
\E[ (V^L_{k , j_{L-1}^{(1)}})^2 ]  \cdots \E[  (V^1_{j_1^{(1)},i})^2] \leq N^{-m}.
 \]
The thesis follows by simple combinatorics. 
\end{proof} 

\begin{proof}[Proof of Proposition \ref{Prop:lazy}]
When $L(N)/N \to 0$, by Lemma \ref{conv.of.diag.elem}
\[
(\bar V^{L,N}_{1,1},\dots,\bar V^{L,N}_{D,D})
 \stackrel{P}{\to}  (1,\dots,1).
\] 
From Lemma \ref{Lemma1}  the off-diagonal elements converge to zero in probability 
since $\Var(\bar V^{L,N}_{k,j})$ converges to $0$ for $k>j$.
The second part of the proof follows combining  Proposition \ref{prop1} and the continuous mapping theorem. 
\end{proof}

\subsection{Skorohod representation}

With reference to \eqref{def.Bartlett.matrix}, 
in what follows we shall use a special representation of the random elements  $V_{i,j}^{\ell}=V_{i,j}^{\ell,N}$ with $L \geq i \geq j \geq 1$ and
$\ell=1,\dots,L$.

 Since we  deal only with convergence in distribution, we can choose any 
representation of these random variables. 
For given random  Bartlett matrices   $[V_{i,j}^{\ell,N}]_{i\geq j}$ with $\ell=1,\dots,L(N)$, set 
\begin{equation}\label{defSandZ}
\begin{split}
&S^{(N,r)}_0=0, \quad 
S^{(N,r)}_m= \sum_{\ell=1}^{m} \log(V^{\ell,N}_{r,r})=  \frac{1}{2} \sum_{\ell=1}^{m} \log\Big ( (V^{\ell,N}_{r,r})^2 ) \quad m=1,\dots,L \\
&  
 \text{and} \quad Z^{(N,r)}_{t}:=S^{(N,r)}_{\lfloor Lt \rfloor} \quad t \in [0,1]. \\
\end{split}
 \end{equation}

\begin{proposition}\label{coupling}
Assume \eqref{H_asym}. 
On   a  suitable (complete)  probability space 
$(\Omega,\mathcal{F},P)$,
there is an  array $[V^{\ell,N}]_{\ell=1,\dots, L=L(N),N \geq D+1}$ such that
\begin{itemize}
\item[(A)] for each given $N$,    $V^{1,N},\dots, V^{L,N}$ are independent lower triangular   $D \times D$ random Bartlett  matrices.
In particular, 
$(V_{r,r}^{\ell,N})^2 \sim Gamma((N-r+1)/2,N/2)$,
and  the $V_{r,r}^{\ell,N}$'s are independent 
for  $r=1,\dots,D$ and $\ell=1,\dots,L(N)$ ($N\geq D+1$ fixed);
\item[(B)] for $ 1 \leq j < i \leq D$ and $N \geq D+1$ 
\[
V_{i,j}^{\ell,N}:=     \sqrt{\frac{L}{N}} \Big ( W^{(i,j)}_{\ell/L}-W^{(i,j)}_{(\ell-1)/L} \Big ) 
\]
where $(W^{(i,j)}_t)_{ t \in [0,1]}$ are independent   Brownian motions;
\item[(C)] if $(Z^{(N,r)}_t)_t$ are defined in \eqref{defSandZ},
one has 
 \[
 \sup_{t \in [0,1] }\left  | Z^{(N,r)}_t -\left( \sqrt{\frac{a}{2}}W^{(r)}_t-\frac{r}{2} a t  \right ) \right |  \stackrel{P}{\to} 0
 \] 
 for $N \to +\infty$, where   $(W_t^{(r)})_{t \in [0,1]}$  are   independent  Brownian motions;
 \item[(D)]  all the Brownian motions  $(W^{(i,j)}_t)_{ t \in [0,1]}$  and $(W_t^{(r)})_{t \in [0,1]}$ with $ D \geq i >j \geq 1$ and $r=1,\dots,D$  are independent. 
\end{itemize} 
\end{proposition}

\begin{proof}

The only nontrivial part of the previous proposition is point (C),  which it is proved below. 
Let $D[0,1]$ be  the space of all functions on $[0, 1]$  that are right-continuous with left-hand limits (rcll) equipped with the sup norm $\|X\|_\infty=\sup_{ t \in [0,1]}\| X_t\|$ and the $\sigma$-field  generated by all evaluation maps
$X \mapsto X_t$. 

For any   $r=1,\dots,D$, 
let  $[\tilde G^N_{\ell,r}]_{\ell,N}$ be an array of  random variables such that $\tilde G^{N}_{\ell,r} \sim Gamma((N-r+1)/2,N/2)$, 
with $[\tilde G^N_{\ell,r}]_{\ell,N}$ independent in $r$ and $\ell$ for any $N$.   Set
\begin{equation}\label{defSandZbis}
\begin{split}
&\tilde S^{(N,r)}_0=0, \quad 
\tilde S^{(N,r)}_m= \sum_{\ell=1}^{m} \frac{1}{2} \log(\tilde G^{\ell}_{N,r}) \quad m=1,\dots,L \\
&  
 \text{and} \quad \tilde Z^{(N,r)}_{t}:=\tilde S^{(N,r)}_{\lfloor Lt \rfloor} \quad t \in [0,1]. \\
\end{split}
 \end{equation}
  If $X=\log(G)$ where $G \sim Gamma(\alpha,\beta)$,  then 
 \begin{equation}\label{meanlogGamma}
\E[X]=\Psi(\alpha)-\log(\beta), 
\end{equation}
where $\Psi(x)=\frac{d}{dx} \log(\Gamma(x))$ is the digamma function. 
By \eqref{meanlogGamma}
\[
\begin{split}
\mu_t^{(N,r)}& =\E \Big [\tilde S^{(N,r)}_{\lfloor Lt \rfloor} \Big ]=
\E \Big [\tilde Z^{(N,r)}_{t} \Big ]
\\
& 
=\frac{1}2{} \frac{\lfloor Lt \rfloor}{L} \frac{L}{N} 
N \big (\Psi( (N-r+1)/2)-\ln(N/2) \big ). \\
\end{split}
\]
Using the asymptotic $\Psi(x)=\ln(x)-(2x)^{-1}+O(x^{-2})$, the fact that  $\frac{\lfloor Lt \rfloor}{L} \to  t$
and $L/N \to a$, one gets 
\begin{equation}\label{conv.moment1}
\lim_{N \to \infty} \sup_{t \in [0,1]}  |\mu_t^{(N,r)} + \frac{t}{2}ar|=0.  
\end{equation}
The sequence of processes 
$(\tilde S^{(N,r)}_{0,\lfloor Lt \rfloor})_{t \in [0,1]}=(\tilde S^{(N,r)}_{\lfloor Lt \rfloor}-\mu_t^{(N,r)})_{t \in [0,1]}$
is a sequence of random walk with zero means.  
Noticing that 
$\log(\bar V^{L,N}_{r,r}) \stackrel{\CL}{=} \tilde S^{(N,r)}_{ L }=\tilde Z^{(N,r)}_{1}$, 
if \eqref{H_asym} holds, then Lemma \ref{conv.of.diag.elem} 
 yields that 
  \[
 \tilde S^{(N,r)}_{ L } \stackrel{\CL}{\to}  Z_r,
 \]
 where $Z_r \sim \mathcal{N}(-a r/2,a/2)$.
Combining this fact with  \eqref{conv.moment1},
one has that  $\tilde S^{(N,r)}_{ 0,L}$
converges in law to a $\mathcal{N}(0,a/2)$. 
 By a suitable version of the  
strong approximation of random walks (see,  e.g, Thm. 12.20 in 
\cite{KallenbergOLD})
one can build 
a probability space $(\Omega,{\mathcal{F}},P)$ and a sequence 
of rcll processes  $(Z_{0,t}^{(r)})_{t \in [0,1]}$ defined on   $(\Omega,{\mathcal{F}},P)$ with the same law of 
$(\tilde S^{(L)}_{0,\lfloor Lt \rfloor})_{t \in [0,1]}$ such that 
\begin{equation}\label{conf_unif_0}
\sup_{ t \in [0,1]} |Z_{0,t}^{(r)}-\sqrt{\frac{a}{2}}W_t^{(r)}|  \to 0  \quad \text{in probability},
\end{equation}
where $(W_t^{(r)})_{t \in [0,1]}$ is a standard Brownian motion 
on $[0,1]$ defined on   $(\Omega,{\mathcal{F}},P)$. 
This construction can be done on the same 
$(\Omega,{\mathcal{F}},P)$ independently for all $r=1,\dots,D$, 
in such a way that $(Z_{0,t}^{(r)}, W_t^{(r)})_t$ are independent. 
Defining  $Z^{(N,r)}_t =Z_{0,t}^{(r)} + \mu_t^{(N,r)} $, then  $(Z^{(N,r)}_t)_{t \in [0,1]}$  has the same law of  the process 
 $(\tilde Z^{(N,r)}_t)_{t \in [0,1]}$ in $D[0,1]$. 
 Combining  
\eqref{conv.moment1} and \eqref{conf_unif_0} one gets also 
 $\sup_{t \in [0,1] } |Z^{(N,r)}_t -\big( \sqrt{\frac{a}{2}}W^{(r)}_t-\frac{r}{2} a t  \big )|  \to 0$ in probability
  for $N \to +\infty$. 
To conclude it suffices to define 
\[
V^{N,\ell}_{r,r} = \exp( ( Z^{(N,r)}_{\ell/L} - Z^{(N,r)}_{(\ell-1)/L})  \qquad \ell=1,\dots,L.
\]
Indeed, one has that, for any $N \geq  D+1$,   $[V^{N,\ell}_{r,r} ]_{\ell,r}$ are independent 
since $(Z^{(N,r)}_t)_t$ and  $(\tilde Z^{(N,r)}_t)$ have the same law 
and  $\tilde Z^{(N,r)}_{\ell/L} -\tilde Z^{(N,r)}_{(\ell-1)/L}$ are independent fo $\ell=1,\dots,L$. Moreover,   
\[
(V^{N,\ell}_{r,r} )^2\stackrel{\CL}{=}\exp( 2( \tilde Z^{(N,r)}_{\ell/L} -\tilde Z^{(N,r)}_{(\ell-1)/L})) =\tilde G^{N}_{\ell,r} \sim Gamma((N-r+1)/2,N/2),
\] and this concludes the proof of (C). 
\end{proof}

\subsection{Filtrations}\label{Sec:filtrations} 
{
In what follows we shall need to use  the processes $(W^{(i,j)}_t)_t$ appearing in  
(B) of  Proposition \ref{coupling}  as stochastic 
integrator, that is 
we shall need to define  $\int_0^t X_s dW^{(i,j)}_s$  with $0 \leq t \leq 1$
for suitable processes  $(X_t)_{t \in [0,1]}$. 
One case has already been encountered,  see  the definition 
of $H({\br})$ in  \eqref{multiplestochasticintegral}. 
To define 
the  stochastic integral we  need to specify a 
 filtration and check the relative measurability conditions for the process $(X_t)_{t \in [0,1]}$. In point of fact the integrands $(X_t)_{t \in [0,1]}$
will depend on  the variables $V^{N,\ell}_{r,r}$, the 
other Brownian motions $(W^{(r)}_t)_t$ and some of 
the $(W^{(i',j')}_t)_{t}$ for $(i',j')\not=(i,j)$ in a specific way, again see  \eqref{multiplestochasticintegral}. 
In order to guarantee that the processes $(X_t)_{t}$ are progressively measurable
All the processes we shall need to integrate will be continuous or   right (left)- continuous, hence 
in order to check that they are progressively measurable it will be enough that they are adapted, see
 Lemma \ref{prog.meas}. 
To this end   
we need to build in an appropriate way suitable  right continuous and complete
filtrations $\bar \CF_{t}^{(i,j)}$.  

In what follows, if $\CH_i$ with $i \in I$ are $\sigma$-fileds we 
denote by $ \vee_{i \in I} \CH_i$ the smallest $\sigma$-filed
which contains $\cup_{i \in I}  \CH_i$.
For any filtration  $(\mathcal{H}_t)_{t\geq 0}$, we denote by  $\mathcal{H}_\infty=\vee_{t \geq 0} \mathcal{H}_t$ 
the $\sigma$-field generated by $(\mathcal{H}_t)_{t\geq 0}$. 
Moreover, if $(W_t)_t$ is a Brownian motion we denote by 
$(\CF^{W}_t)_t$ its natural filtration. 

Define
\[
\CH_0=\sigma \big (V^{N,\ell}_{r,r} :   \ell=1,\dots,L(N),  r=1,\dots,D, N > D\big) \bigvee  
\Big (\bigvee_{r=1}^D \CF^{W^{(r)}}_\infty \Big ).
\]
For $r=D-1,\dots,1$ define
\[
\CF_t^{(D,r)}=  \CH_0 \bigvee \CF^{W^{(D,r)}}_t,
\qquad 
\CN^{(D,r)}=\{ N \subset \Omega: \exists A \in \CF_\infty^{(D,r)}: N \subset A  : P(A)=0\}
\]
and 
\[
\bar\CF_t^{(D,r)}=\CF_t^{(D,r)}  \bigvee  \CN^{(D,r)}.
\]
Iteratively, for $r_2=D-1,\dots,2$ and, given $r_2$,  for $1 \leq r_1 < r_2 \leq D-1$, define
\[
\CG^{(r_2,r_1)}= \bigvee_{r_2 \leq q<p \leq D}  \bar \CF^{(p,q)}_\infty 
\supset \CG_0 ,
\qquad 
\CF_t^{(r_2,r_1)}=\CF^{W^{(r_2,r_1)}}_t \bigvee \CG^{(r_2,r_1)},
\]
\[
\CN^{(r_2,r_1)}=\{ N \subset \Omega: \exists A \in \CF_\infty^{(r_2,r_1)}: N \subset A  : P(A)=0\}
\]
and 
\[
\bar\CF_t^{(r_2,r_1)}=\CF_t^{(r_2,r_1)}  \bigvee  \CN^{(r_2,r_1)}.
\]

Using  Lemma \ref{LemmaP1} and Lemma \ref{prog.meas}  one obtains the next result. 

\begin{lemma}\label{lemma_wellposedness} 
For any $1 \leq r_1 < r_2 \leq D$, $\big ( (W_t^{(r_2,r_1)})_{t \in [0,1]}, 
(\bar \CF_t^{(r_2,r_1)})_{t \in [0,1]})$ are standard Brownian motions. Moreover, 
all the integrals in  \eqref{multiplestochasticintegral} are well-defined. 
\end{lemma}

}

\subsection{Proof of Proposition \ref{LNlimitwithKgendim}}
We start from re-writing in a more convenient way 
  \eqref{barVkjA}.
To this end, if  $\mathbf{j}=(j_0, j_1 \dots,  j_L) $ is a vector of integers such that  
$j_0:=i \leq j_1 \dots \leq j_L   = k $,  define  $h=h(\mathbf{j})$ to be the number of 
$\ell=1,\dots,L$ such that  $j_{\ell}\not=j_{\ell-1}$  and  let
\[
r_0:=i  < r_1 < \cdots < r_{h-1} <r_h:=k
\]
be the $h+1$  distinct values in $\mathbf{j}=(j_0,\dots,j_{L})$.  
For $s=0,\dots,h$ set  also
\[
m_s=\#\{ \ell=1,\dots,L : (j_{\ell},j_{\ell-1})=(r_s,r_s)\}. 
\]
Note that $m_0+m_1+\dots m_h=L-h$. 
Finally set $n_0=0$ and for $j = 1,\dots,h+1$
\[
n_j=\sum_{s=0}^{j-1} m_s + j =n_{j-1}+m_{j-1}+1
\]
in other words, $n_0=0, n_1=m_0+1, n_2=m_0+m_1+2, \dots$
Note that $n_h=L-m_h$ and $n_{h+1}=L+1$.  
For example, let $L=5,i=1,k=4$ and $(j_0,j_1,j_2,j_3,j_4,j_5)= (1,1,3,3,3,4)$. 
This corresponds to 
\[
 V^1_{1,1} V^2_{3,1} V^3_{3,3}V^4_{3,3}  V^{5}_{4,3}  
\]
and here $h=2, \,\, r_0=1<r_1=3<r_2=4, \,\, m_0=1, m_1=2 , m_2=0$.

With this notation one can write 
\[
\begin{split}
V^1_{j_{1},j_0}  \dots  V^{L}_{j_L,j_{L-1}}  
& = \Big( \prod_{j=1}^h  \Big ( 
\prod_{\ell=n_{j-1}+1}^{n_j-1} V_{r_{j-1},r_{j-1}}^{\ell}  \Big ) V_{r_j,r_{j-1}}^{n_j} \Big)  \prod_{\ell=n_h+1}^{L}  V_{r_h,r_h}^{\ell}  \\
& =  \Big (  \prod_{j=1}^{h+1} 
\prod_{\ell=n_{j-1}+1}^{n_j-1} V_{r_{j-1},r_{j-1}}^{\ell}  \Big ) 
 \Big (  \prod_{j=1}^{h} 
V_{r_j,r_{j-1}}^{n_j} \Big )  \\
\end{split}
\]
Recalling   \eqref{defSandZ}, one gets %
\[
\begin{split}
 V^1_{j_{1},j_0}  \dots  V^{L}_{j_L,j_{L-1}}  
 & = e^{ \sum_{j=1}^{h+1} 
\sum_{\ell=n_{j-1}+1}^{n_j-1} \frac{1}{2}   \log( (V_{r,r}^{\ell})^2 ) }  \prod_{j=1}^{h}  V_{r_j,r_{j-1}}^{n_j}   \\
& 
=e^{ \sum_{j=0}^{h} 
\big (S^{(N,r_{j})}_{n_{j+1}-1}-S^{(N,r_{j})}_{n_{j}} \big )  
}   \prod_{j=1}^{h}  V_{r_j,r_{j-1}}^{n_j} 
\\
&
=e^{S_L^{(N,r_{h})}+ \sum_{j=1}^{h} 
\big (S^{(N,r_{j-1})}_{n_{j}-1}-S^{(N,r_{j})}_{n_{j}} \big )  
}     \prod_{j=1}^{h}  V_{r_j,r_{j-1}}^{n_j} 
\\
&= \Big ( \frac{L}{N} \Big)^{\frac{h}{2}}
e^{ Z^{(N,r_{h})}_{1} + \sum_{j=1}^{h} 
\big (Z^{(N,r_{j-1})}_{(n_{j}-1)/L}-Z^{(N,r_{j})}_{n_{j}/L} \big )  
}   \prod_{j=1}^{h}V_{r_j,r_{j-1}}^{n_j} \\
& 
= \Big ( \frac{L}{N} \Big)^{\frac{h}{2}}
e^{ Z^{(N,r_{h})}_{1}  }  \prod_{j=1}^{h} 
e^{ 
\big (Z^{(N,r_{j-1})}_{(n_{j}-1)/L}-Z^{(N,r_{j})}_{n_{j}/L} \big )  
} 
\Big ( W^{(r_j,r_{j-1})}_{n_j/L}-W^{(r_j,r_{j-1})}_{(n_j-1)/L} \Big).
\\
\end{split}
\]
where in the last equality we used the fact that  
\[
V_{r_j,r_{j-1}}^{n_j}=
\sqrt{\frac{L}{N}} \Big ( W^{(r_j,r_{j-1})}_{n_j/L}-W^{(r_j,r_{j-1})}_{(n_j-1)/L} \Big ) 
\]
where  $(W^{(r_j,r_{j-1})}_t)_{t \in [0,1]}$ are the independent standard Brownian motions of Proposition \ref{coupling}.

Setting $\CR_{k,i}^h=\{ \br=(r_0,\dots,r_h):r_0:=i<r_1<\cdots<r_{h-1}<r_h:=k\}$, 
 $\CN_{L,h} =\{\bnn=(n_1,\dots,n_h) \in \N^h : 1 \leq n_1<n_2\dots<n_h \leq L\}$
and 
\[
{I}^L(r_j,n_j) =
e^{ 
\big (Z^{(N,r_{j-1})}_{(n_{j}-1)/L}-Z^{(N,r_{j})}_{n_{j}/L} \big )  
} 
\Big ( W^{(r_j,r_{j-1})}_{n_j/L}-W^{(r_j,r_{j-1})}_{(n_j-1)/L} \Big)
\]
one can write 
\[
\bar V^{L,N}_{k,i} = 
\sum_{i \leq  j_1 \leq  \dots \leq j_{L-1} \leq k}
 V^1_{j_{1},i} V^1_{j_{2},j_1} \cdots
 V^{L}_{k, j_{L-1}}  =
 \sum_{h=1}^{k-i} \sum_{ \br \in \CR_{k,i}^h }   \Big ( \frac{L}{N} \Big)^{\frac{h}{2}}
e^{ Z^{(N,r_{h})}_{1}  } 
\sum_{\bnn \in \CN_{L,h}}  \prod_{j=1}^{h}  {I}^L(r_j,n_j).  \\
\]
We now consider the inner part of the sum:   
\[
\begin{split}
H_L(\br)  & :=   \Big ( \frac{L}{N} \Big)^{\frac{h}{2}} e^{ Z^{(N,r_{h})}_{1}  } 
  \sum_{ \br \in \CR_{k,i}^h }
\sum_{\bnn \in \CN_{L,h}}  \prod_{j=1}^{h}  {I}^L(r_j,n_j)  
 \\
& =
  \Big ( \frac{L}{N} \Big)^{\frac{h}{2}}
e^{ Z^{(N,r_{h})}_{1}  } 
\sum_{n_1=1}^{L-(h-1)}  {I}^L(r_1,n_1)  \sum_{n_2=n_1+1}^{L-(h-2)}  {I}^L(r_2,n_2) 
\dots  \sum_{n_h=n_{h-1}+1}^{L}  
 {I}^L(r_h,n_h) .  \\
 \end{split}
\]

It is plain to check that 
\[
H_L(\br) = \Big ( \frac{L}{N} \Big)^{\frac{h}{2}} e^{ Z^{(N,r_{h})}_{1}  } 
\Psi^L_1(0;\br)
\]
where, given  $\br=(r_1,\dots,r_h)$, the functions  $\Psi^L_j(n_{j-1};\br)$
for $j=1,\dots,h$ 
 are defined by the backward recursion: 
\[
\begin{split}
& \text{for $n_{h-1} = {h-1}, \dots, L-1$}   \\
& \qquad  \Psi^L_h(n_{h-1};\br) = 
 \sum_{n_h=n_{h-1}+{1}}^{L}  {I}^L(r_h,n_h)   \\
& \text{for $j=h-1,\dots,2$  and  $ n_{j-1} = {j-1}, \dots, L-{(h-j)-1}$} \\
& \qquad   \Psi^L_j(n_{j-1};\br) =
\sum_{n_j=n_{j-1}+1}^{L-(h-j)}  {I}^L(r_j,n_j)  \Psi^L_{j+1}(n_j;\br)  
\\
& \text{and} \\
& \qquad   \Psi^L_1(0;\br) =
\sum_{n_1=1}^{L-(h-1)}  {I}^L(r_1,n_1)  \Psi^L_{2}(n_1)  .
\\
\end{split} 
\]
We now  express $H_L(\br)$ as a multiple stochastic integral. To this end 
define 
\[
g_{j ,\br,N}(t):= e^{Z^{(N,r_{j-1})}_{t}- Z^{(N,r_j)}_{(t+1/L) \wedge 1}} \quad (j=1,\dots,h). 
 \]
 Note that
 \[
  {I}^L(r_h,n_h)  =
  e^{ 
\big (Z^{(N,r_{h-1})}_{(n_{h}-1)/L}-Z^{(N,r_{h})}_{n_{h}/L} \big )  
} 
\Big ( W^{(r_h,r_{h-1})}_{n_j/L}-W^{(r_h,r_{h-1})}_{(n_j-1)/L} \Big)=
  \int_{(n_{h}-1)/L}^{n_{h}/L}   g_{h,\br,N}(t_{h})dW^{(r_{h},r_{h-1})}_{t_{h}}.
 \]
Hence, if  
\[
\mathcal{G}_{h,\br,N}(\tau):=\int_{\tau\wedge 1}^1   g_{h,\br,N}(t_h)  dW^{(r_{h},r_{h-1})}_{t_h} 
\]
then
\[
\Psi^L_h(n_{h-1};\br) =\
 \sum_{n_h=n_{h-1}+{1}}^{L} \int_{(n_{h}-1)/L}^{n_{h}/L}   g_{h,\br,N}(t_{h})dW^{(r_{h},r_{h-1})}_{t_{h}} =\CG_{h,\br,N}(n_{h-1}/L).
\]
Using this relation one can write 
\[
\begin{split}
 {I}^L(r_{h-1},n_{h-1}) 
\Psi^L_{h}(n_{h-1};\br)& = \int_{(n_{h-1}-1)/L}^{n_{h-1}/L}   g_{h-1,\br,N}(t_{h-1})  \CG_{h,\br,N}(n_{h-1}/L)   dW^{(r_{h-1},r_{h-2})}_{t_{h-1}}  \\
& 
= \int_{(n_{h-1}-1)/L}^{n_{h-1}/L}   g_{h-1,\br,N}(t_{h-1})  \CG_{h,\br, N}((\lfloor Lt_{h-1} \rfloor +1)/L )   dW^{(r_{h-1},r_{h-2})}_{t_{h-1}}  .
\\
\end{split} 
\]
Hence, taking the sum as above, one gets
\[
\begin{split}
\Psi^L_{h-1}(n_{h-2};\br)  & =   \int_{n_{h-2}/L}^{1-1/L}   g_{h-1,\br,N}(t_{h-1})  \CG_{h,\br,N}((\lfloor Lt_{h-1} \rfloor +1)/L )   dW^{(r_{h-1},r_{h-2})}_{t_{h-1}} \\
 & =   \int_{n_{h-2}/L}^{1}   \mathbb{I} \{ t_{h-1} \leq 1-1/L \} g_{h-1,\br,N}(t_{h-1})  \CG_{h,\br, N}((\lfloor Lt_{h-1} \rfloor +1)/L )   dW^{(r_{h-1},r_{h-2})}_{t_{h-1}} .\\
\\
\end{split}
\]
So that, 
defining 
\[
\CG_{h-1,\br,N}(\tau  ):= \int_{\tau \wedge 1}^{1} \mathbb{I}  \{ t_{h-1} \leq 1-1/L \}    g_{h-1,\br,N}(t_{h-1})  \CG_{h,\br, N}\Big (\frac{\lfloor Lt_{h-1} \rfloor +1}{L} 
  \Big )   dW^{(r_{h-1},r_{h-2})}_{t_{h-1}} ,
\]
one obtains 
\[
\Psi^L_{h-1}(n_{h-2};\br)  = \CG_{h-1,\br,N}(n_{h-2}/L ). 
\]
Iterating this construction 
\[
\Psi^L_{j}(n_{j-1};\br) = \CG_{j,\br,N}(n_{j-1}/L ) 
\]
where 
\[
\CG_{j,\br,N}(\tau)=   \int_{\tau \wedge 1}^1 \mathbb{I}\{t_{j} \leq 1-(h-j)/L\}   g_{j,\br,N}(t_{j})  \CG_{j+1,\br,N}\Big (\frac{\lfloor Lt_{j} \rfloor +)}{L} \Big )   dW^{(r_{j},r_{j-1})}_{t_{j}} .
\]

Note that all the stochastic integrals aboved are well-defined with respect to 
the right-continuous and complete  filtrations $(\bar \CF_t^{((r_{j},r_{j-1}))})_{t \in [0,1]}$ introduced in  Subsection \ref{Sec:filtrations}. 
Let $(W^{(r)}_t)_{t \in [0,1]}$ for $r=1,\dots,D$ be the Brownian motions of Proposition \ref{coupling}, 
recall that by \eqref{def.Zkt} $Z^{(r)}_t:=\sqrt{\frac{a}{2}}W^{(r)}_t-\frac{a r}{2}  t$ and set   
for $t \in [0,1]$
\[
g_{j ,\br}(t)=e^{Z^{(r_{j-1})}_{t}- Z^{(r_j)}_{t}}.
\]

\begin{lemma}
For every $1 \leq i < k \leq D$, every $h=1,\dots, k-i$,  every  $\br  \in \CR_{k,i}^h$
and $j=1,\dots,h$
\begin{equation}\label{convzGeneral}
  \sup_{t \in [0,1]} | g_{j ,\br,N}(t)-g_{j ,\br}(t) | \stackrel{P}{\to} 0 
 \end{equation}
\end{lemma}

\begin{proof}
Observe that 
\[
\begin{split}
& R_N:= \sup_{t \in [0,1]} \Big|   Z^{(N,r_{j-1})}_{t}  -Z^{(N,r_{j})}_{(t+\frac{1}{L}) \wedge 1 } -Z^{(r_{j-1})}_{t}  -Z^{(r_{j})}_{t} \Big |  
 \leq \sup_{t \in [0,1]}  \Big|     Z_{t}^{(N,r_{j-1})}
-\sqrt{\frac{a}{2}}W^{(r_{j-1})}_{t}+\frac{ar_{j-1} t}{2} \Big|  \\
&  +   \sup_{t \in [0,1]}  \Big|  
\sqrt{\frac{a}{2}}W^{(r_{j})}_{(t+\frac{1}{L}) \wedge 1}-\frac{ar_{j} }{2}  \big( (t+ \! \frac{1}{L}) \wedge 1\big)
-\sqrt{\frac{a}{2}}W^{(r_{j})}_{t}+\frac{ar_{j} t }{2}  
\Big|
\\
&\qquad   + \sup_{t \in [0,1] }  \Big| Z_{ (t+\frac{1}{L}) \wedge 1 }^{(N,r_j)}-
\sqrt{\frac{a}{2}}W^{(r_{j})}_{(t+\frac{1}{L}) \wedge 1}+\frac{ar_{j} }{2}  \big( (t+ \! \frac{1}{L}) \wedge 1\big) \Big|  \Big)\\
\\
\end{split}
\]
Proposition  \ref{coupling} (C) and the continuity of $(W^{(r_j)}_{t})_t$ yields that   
\[
 \sup_{t \in [0,1]} \Big|   Z^{(N,r_{j-1})}_{t}  -Z^{(N,r_{j})}_{(t+\frac{1}{L}) \wedge 1 } -Z^{(r_{j-1})}_{t}  -Z^{(r_{j})}_{t} \Big |   \stackrel{P}{\to} 0
\]
and hence $R_N \to 0$ in probability. 
For any two bounded functions $a_t$ and $b_t$ 
\[
|e^{a_t}-e^{b_t} | \leq 
\|  a-b\|_\infty e^{\| a  -b\|_\infty+
\|b\|_\infty}.
\]
Hence $\|g_{j ,\br,N}-g_{j ,\br}\| \leq R_N e^{R_N+ \| g_{j ,\br}\|}$ but $ R_N e^{R_N+ \| g_{j ,\br}\|} \to 0$ in probability. 
\end{proof}

Since $g_{h ,\br}(t)=e^{Z^{(r_{h-1})}_{t}- Z^{(r_h)}_{t}}$ is continuous and $\bar \CF^{(r_{h},r_{h-1})}_0$ measurable, one can define 
\[
\CG_{h,\br}(\tau):=\int_{\tau}^1 g_{h ,\br}(t) \  dW^{(r_{h},r_{h-1})}_{t_h}   \qquad \tau \in [0,1].
\]
Note that  $(\CG_{h,\br}(\tau))_{\tau \in [0,1]}$ turns out to be $(\bar \CF^{(r_{h},r_{h-1})}_\tau)_{\tau \in [0,1]}$ adapted and continuous 
and 
\[
(\bar \CF^{(r_{h},r_{h-1})}_\tau)_{\tau \in [0,1]} \subset \bar \CF^{(r_{h-1},r_{h-2})}_0.
\] Hence, one can 
recursively define  for $j=h-1,\dots,1$ 
\[
\CG_{j,\br}(\tau)=   \int_{\tau}^1g_{j ,\br}(t) \CG_{j+1,\br}(t_j )   dW^{(r_{j},r_{j-1})}_{t_{j}}  \qquad \tau \in [0,1]. 
\]
Note also that all these processes are continuous. 
Using  \eqref{convzGeneral} and Lemma \ref{Lemma.conv.of.integral} (or directly Lemma \ref{Lemma_approx_int_W}), it follows that 
$\sup_{\tau \in [0,1]} | \CG_{h,\br,N}(\tau) - \CG_{h,\br}(\tau)|  \stackrel{P}{\to} 0$. 
Using this result and again    \eqref{convzGeneral} and Lemma \ref{Lemma.conv.of.integral}, one obtains 
$\sup_{\tau \in [0,1]} | \CG_{h-1,\br,N}(\tau) - \CG_{h-1,\br}(\tau)|  \stackrel{P}{\to} 0$. 
Iterating, 
\[
\sup_{\tau \in [0,1]} | \CG_{1,\br,N}(\tau) - \CG_{1,\br}(\tau)|  \stackrel{P}{\to} 0
\]
so that $\CG_{1,\br,N}(0) \stackrel{P}{\to}  \CG_{1,\br}(0)$. Which proves that 
\[
H_L(\br) =\Big ( \frac{L}{N} \Big)^{\frac{h}{2}} e^{ Z^{(N,r_{h})}_{1}  }  \CG_{1,\br,N}(0)  \stackrel{P}{\to}  a^{\frac{h}{2}}  e^{ Z^{(r_{h})}_{1}  } 
 G_{1,\br}(0)=H(\br)
\]
where $H(\br)$ has been defined in \eqref{multiplestochasticintegral}. 
Proposition \eqref{LNlimitwithKgendim} follows easily.

\subsection*{ Acknowledgements and Disclosure of Funding}

{F.B. is partially supported by the MUR - PRIN project ``Discrete random structures for Bayesian learning and prediction'' no. 2022CLTYP4.
AP.R. is supported by $\#$NEXTGENERATIONEU (NGEU) and funded by the Ministry of University and Research (MUR), National Recovery and Resilience Plan (NRRP), project MNESYS (PE0000006) ``A Multiscale integrated approach to the study of the nervous system in health and disease'' (DN. 1553 11.10.2022).
}

\appendix
\section*{Appendix} 
\addcontentsline{toc}{section}{Appendix} 

\section{Matrix Normal distribution}\label{appendixA}

 A random matrix 
$Z$ of dimension $n_1 \times n_2$  has a centred matrix normal distribution 
with parameters $(\Sigma_1,\Sigma_2)$ (with $\Sigma_i$'s positive symmetric $n_i \times n_i$ matrices),
if for any  matrix $S$ of dimension $n_2 \times n_1$ 
\begin{equation}\label{PMN1}
\E[e^{i \tr(S Z)}]=\exp\left \{-\frac{1}{2} \tr(S\Sigma_1 S^\top \Sigma_2) \right\}.
\end{equation}
In symbols $Z \sim \mathcal{MN}(0,\Sigma_1,\Sigma_2)$. 

Two useful properties are reported here, see \cite{Gupta2000} for details.

\begin{enumerate}[label={(P\arabic*)},start=1]

\item {\it  Linear transformation of matrix normals. }  \label{AA}
 Given two matrices $H$ and $K$ with compatible shape, if $Z \sim \mathcal{MN}(0,\Sigma_1,\Sigma_2)$
then 
\[
H Z K \sim \mathcal{MN}(0,H\Sigma_1H^\top,K^\top\Sigma_2K).
\]
 \item  \label{BB}
{\it    Equivalence with the multivariate normal. }
  $Z \sim \mathcal{MN}(0,\Sigma_1,\Sigma_2)$ if and only if 
$\vvec(Z) \sim \CN(0,\Sigma_2\otimes \Sigma_1)$
\end{enumerate}

\section{Few results on stochastic integrals}

In what follows let be  $(\CF_t)_{t \geq 0}$ be a complete and rigth-continuous  filtration  on $(\Omega,\CF,P)$
and $(W_t,\CF_t)_{t \geq 0} $ is a Browninan motion on $(\Omega,\CF,P)$ with respect to the  filtration $(\CF_t)_{t \geq 0}$.


\begin{lemma}[Prop. 1.3 \cite{KaratzasShreve}]\label{prog.meas}
Let $(X_t)_{t \geq 0}$ a  $(\CF_t)_t$ adapted process. If it is left continuous or right continuous, then  $(X_t)_{t \geq 0}$
is    $(\CF_t)_{t \geq 0}$-progressively measurable. 
\end{lemma}

\begin{lemma}\label{Lemma_approx_int_W}
 Let $(X_t^\infty)_t$ be a $(\CF_t)_t$   progressively measurable process and 
 $(X_t^N)_t$ $N=1,\dots,+\infty$ be a sequence of $(\CF_t)_{t \geq 0}$ progressively measurable processes such  that 
 $P\{ \int_0^T (X_t^N)^2 dt<+\infty\}=1$ 
 for some for $T<+\infty$  and  for  every $N \leq +\infty$. 
If $\int_0^T (X_t^N-X_t^{\infty})^2 dt \stackrel{P}{\to} 0$ then 
\[
\sup_{t \in [0,T]}\Big  |\int_0^t X_s^NdW_s- \int_0^t X_s^\infty dW_s \Big|  \stackrel{P}{\to} 0. 
\]
\end{lemma}

See Proposition 2.26 in Chp. 3 of \cite{KaratzasShreve}.  

%

\begin{lemma}\label{Lemma.conv.of.integral} Let   $(W_t,\CF_t)_{t  \in [0,1]}$ is a standard  Browninan motion on $(\Omega,\CF,P)$ with respect to the  filtration $(\CF_t)_{t  \in [0,1]} $.
Let $(X^{k,N}(t))_{t  \in [0,1]}$ be a $(\CF_t)_{t  \in [0,1]} $ progressively  measurable process ($k=1,2$) 
and $(X^{k,\infty}(t))_{t  \in [0,1]}$  ($k=1,2$) be  $(\CF_t)_{t  \in [0,1]} $   adapted and continuous processes. 
Let $L_N$ and $\epsilon_N$ such that $L_N \to +\infty$  and $\epsilon_N \to 0$ when $N \to \infty$. 
Set $A_N(t)=((\lfloor L_N t \rfloor +1)/L_N )\wedge 1$ and assume that also 
$(X^{2,N}(A_N(t)))_{t  \in [0,1]}$ is $(\CF_t)_{t  \in [0,1]} $ progressively  measurable. 
For $\tau \in [0,1]$ let
\[
G_N(\tau):=\int_{\tau}^1  X^{1,N}(t) X^{2,N}(A_N(t)) \mathbb{I}\{t \leq 1-\epsilon_N\}  dW_t 
\]
and 
\[
G_\infty(\tau):=\int_{\tau}^1  X^{1,\infty}(t) X^{2,\infty} (t) dW_t .
\]
If 
\begin{equation}\label{contozeroXk}
\sup_{ t \in [0,1]} | X^{k,N}(t)-X^{k,\infty}(t)| \stackrel{P}{\to} 0 \quad k=1,2 ,
\end{equation}
 then $G_\infty(\tau)$ is continuous and 
\[
\sup_{ \tau \in [0,1]} | G_N(\tau)-G_\infty(\tau)|  \stackrel{P}{\to}  0 .
\]
\end{lemma}

\begin{proof}
By Lemma \ref{Lemma_approx_int_W} it suffices to show that
\[
\begin{split}
\int_0^1 |  &X^{1,N}(t) X^{2,N}(A_N(t))  \mathbb{I}\{t \leq 1-\epsilon_N\}-  X^{1,\infty}(t)X^{2,\infty}(t)  |^2 dt \\
& =\int_0^{1-\epsilon_N}  |  X^{1,N}(t) X^{2,N}(A_N(t))-  X^{1,\infty}(t)X^{2,\infty}(t)  |^2 dt 
 +\int_{1-\epsilon_N}^1 | X^{1,\infty}(t)X^{2,\infty} (t) |^2 dt \\
\end{split}
\]
converges to zero in probability.
As $N \to +\infty$ the last term $\int_{1-\epsilon_N}^1 | X^{1,\infty}(t)X^{2,\infty} (t) |^2 dt$ converges in probability to zero. 
Moreover, 
\[
\begin{split}
 \int_0^{1-\epsilon_N} &  |  X^{1,N}(t) X^{2,N}(A_N(t))-  X^{1,\infty}(t) X^{2,\infty}(t)  |^2 dt  \\
& 
 \leq  4 \Big ( \int_0^{1-\epsilon_N}  |  X^{1,N}(t)  \Big (X^{2,N}(A_N(t))- X^{2,\infty}(A_N(t))\Big) |^2dt \\
&\quad +  \int_0^{1-\epsilon_N}    |  \Big(X^{1,N}(t)-  X_t^{1,\infty}\Big) X^{2,\infty}(A_N(t))  |^2dt \\
& \quad +  \int_0^{1-\epsilon_N}  \!\!\!  | X^{1,\infty}(t)  \Big ( X^{2,\infty}(A_N(t)) -
  X_{t}^{2,\infty} (t) \Big )|^2 dt \Big)  \\
&     \leq  4   
   \|   X^{1,N}\|_\infty^2     \|   X^{2,N} -X^{2,\infty} \|_\infty^2 + 
 4   \|   X^{2,N}\|_\infty^2     \|   X^{1,N} -X^{1,\infty} \|_\infty^2 \\
    & 
  \quad+
    4  \|   X^{1,\infty}\|_\infty^2    \sup_{t \in [0,1]} |  X^{2,\infty}(A_N(t)) -
  X_{t}^{2,\infty} (t)|^2 . 
   \\
\end{split}
\]
Now $  \|   X^{1,\infty}\|_\infty^2    \sup_{t \in [0,1]} |  X^{2,\infty}(A_N(t)) -
  X_{t}^{2,\infty} (t)|^2 \to 0$ a.s. by continuity of $X^{2,\infty}$, moreover
$ \|   X^{k,N}\|_\infty  \leq \|   X^{k,N}- X^{k,\infty} \|_\infty+ \| X^{k,\infty} \|_\infty \stackrel{P}{\to}  \| X^{k,\infty} \|_\infty $ 
by \eqref{contozeroXk}, so that again by  \eqref{contozeroXk}
\[
     \| X^{1,N}\|_\infty^2     \|   X^{2,N} -X^{2,\infty} \|_\infty^2 + 
    \| X^{2,N}\|_\infty^2     \|   X^{1,N} -X^{1,\infty} \|_\infty^2  \stackrel{P}{\to} 0 .
\]
\end{proof}

If $(W_t,\CF^W_t)_{t \geq 0}$ is a Brownian motion on a complete probability space $(\Omega,\CF,P)$, being   $(\CF^W_t)_{t \geq 0}$  the natural filtration, 
one can extend $(\CF^W_t)_{t \geq 0}$ to a complete and  right-continuous (actually continuous)  $(\bar \CF_t)_{t \geq 0}$ filtration in such a way that 
$(W_t,\bar \CF_t)_t$  is still a Brownian motion. To this end  it suffices to consider 
\[
\bar \CF_t= \sigma(\CF^W_t \cup \CN) 
\] 
where $\CN=\{ N \subset \Omega: \exists A \in \CF_\infty^W: N \subset A  : P(A)=0\}$ are the null set in $\CF_\infty^W$. See Chapter 2 Section 7 in \cite{KaratzasShreve}. 

We shall need a slightly more general construction.
The proof of the next result  follows with an argument analogous of the one given in Chapter 2 Section 2.A in  \cite{KaratzasShreve}. 

\begin{lemma}\label{LemmaP1} 
 Let $(W_t,\CF^W_t)_{t \geq 0}$ be a Brownian motion and let  $\CG_0$ be a sub-$\sigma$-field of $\CF$ such that  $\CG_0$ and $\CF_\infty^W$ are independent.
Set $\CG_t=\sigma(\CG_0 \cup \CF^W_t)$, $\CN^*=\{ N \subset \Omega: \exists A \in \CG_\infty: N \subset A  : P(A)=0\}$
and $\bar \CF_t=\sigma(\CG_t  \cup \CN^*)$. Then $(W_t,\bar \CF_t)_{t \geq 0}$ is a standard  Brownian motion, i.e. it  is a Brownian motion and  $(\bar \CF_t)_{t \geq 0}$ is complete and right-continuous. 
\end{lemma}

\section{Posterior of a mixture of normals}\label{AppendixMixture}


Let $\Sigma: \CS^+_{D} \to \CS^+_{D  (P+1)}$ and write 
\[
\Sigma(Q)=\begin{pmatrix}
\Sigma_{00}(Q) &  \Sigma_{01}(Q) \\
\Sigma_{01}^\top  (Q) &  \Sigma_{11}(Q) \\
\end{pmatrix}
\]
where $\Sigma_{00}(Q)$,  $\Sigma_{01}(Q)$, $\Sigma_{11}(Q)$,  
are the 
$D \times D$, $D \times DP  $ and $DP \times DP$ submatrices of $\Sigma(Q) $. 
Let $\mathcal{Q}$ be a probability distribution on $\CS^+_{D}$. 
Assume the following  joint distribution in $(\by_{1:P},\bz_0,\bz_{1:P},Q)$: 
\begin{equation}\label{joint}
\begin{split}
\mu(dQ d\bz_{0:P} d\by_{1:P}) & =
\mu(d\by_{1:p}|\bz_{1:P},Q) \mu(d\bz_{0:P}|Q) \mu(dQ)  \\
& 
:=\CN(d\by_{1:p}| \Sigma_{11}^{1/2} \bz_{1:P},\beta^{-1} \eI_{DP} ) 
\CN(d\bz_{1:P}| \mathbf{0},   \eI_{D(P+1)} )  \mathcal{Q}(dQ), 
\end{split}
\end{equation}
where to simplify the notation we write $\Sigma$ in place of $\Sigma(Q)$. 
Note that above $\mathcal{Q}(dQ)=\mu(dQ)$, $\bz_{1:P}$ and $\by_{1:}$ are in $\RE^{DP}$, $\bz_0 \in \RE^D$ and
$\bz_{0:P}^\top=(\bz_0^\top,\bz_{1:p}^\top)$. 
One has 
\begin{equation}\label{zQ|y}
\mu(d \bz_{0:P} dQ| \by_{0:P} ) =  \mu(d \bz_{0}) \mu(  d \bz_{1:P}   dQ | \by_{1:P})= \mu(d \bz_{0})  \mu(  d \bz_{1:P} | Q, \by_{1:P})\mu(dQ|  \by_{1:P}) 
\end{equation}
and   
\[
\mu(  d \bz_{1:P}dQ | \by_{1:P}) =\mu(  d \bz_{1:P} | Q, \by_{1:P})\mu(dQ|  \by_{1:P})   \propto f(\bz_{1:P},\by_{1:P}| Q )  \mathcal{Q}(dQ) d\bz_{1:P}
\]
with 
\[
f(\bz_{1:P},\by_{1:P}| Q ):=   
{e^{-\frac{\beta}{2} (\Sigma_{11}^{1/2} \bz_{1:P} -\by_{1:P})^\top ( \Sigma_{11}^{1/2} \bz_{1:P}-\by_{1:P})} } 
e^{- \frac{1}{2}\bz_{1:P}^\top \bz_{1:P} } .
\]
Setting  
\[
\mathbf{m}_1= \mathbf{m}_1(Q|\by_{1:P})= \beta ( {\beta}\Sigma_{11}+\eI_{DP} )^{-1} \Sigma_{11}^{1/2} \by_{1:p}, 
\]
which is well-defined since $ {\beta}\Sigma_{11}+\eI_{DP}>0$, 
one checks that  
\[
\begin{split}
& \frac{\beta}{2} (\Sigma_{11}^{1/2} \bz_{1:P}-\by_{1:P})^\top (\Sigma_{11}^{1/2} \bz_{1:P} -\by_{1:P})  - \frac{1}{2}\bz_{1:P}^\top \bz_{1:P}  \\
&=\frac{\beta}{2} \Big [  \by_{1:p} ^\top( \eI_{DP}- \beta  \Sigma_{11}^{1/2}  ({\beta}\Sigma_{11}+\eI_{DP} )^{-1}\Sigma_{11}^{1/2}  ) \by_{1:p}  \Big] \\
& +\frac{1}{2} (\bz_{1:P}-\mathbf{m}_1)^\top ( {\beta} \Sigma_{11}+\eI_{DP})  (\bz_{1:P}-\mathbf{m}_1)  .
\\
\end{split}
\]
Noticing  that 
\[
( \eI_{DP}- \beta  \Sigma_{11}^{1/2}  ({\beta}\Sigma_{11}+\eI_{DP} )^{-1}\Sigma_{11}^{1/2}  )
= ({\beta}\Sigma_{11}+\eI_{DP} )^{-1}
\]
one can write 
\[
f(\bz_{1:P},\by_{1:P}| Q )
={e^{-\frac{1}{2} \Psi (Q|\by_{1:P}) }} 
 \frac{ e^{-\frac{1}{2} (\bz_{1:P}-\mathbf{m}_1)^\top ( {\beta} \Sigma_{11}+\eI_{DP})  (\bz_{1:P}-\mathbf{m}_1)  }}{ \det(( {\beta} \Sigma_{11}+\eI_{DP})
 ^{-1})^{1/2} } 
\]
where
\[
\Psi (Q|\by_{1:P})=
\beta \by_{1:P}^\top ( \eI_{DP}+\beta \Sigma_{11}(Q))^{-1} \by_{1:P}
 +\log(\det(\eI_{DP} +\beta \Sigma_{11}(Q)).
\]
Then  
\begin{equation}\label{cond_Z_givenQy}
\mu(d\bz_{1:P},dQ|\by_{1:P})  =\frac{ e^{-\frac{1}{2} (\bz_{1:P}-\mathbf{m}_1)^\top ( {\beta} \Sigma_{11}+\eI_{DP})  (\bz_{1:P}-\mathbf{m}_1)  }}{(2 \pi)^{\frac{DP}{2}} \det(( {\beta} \Sigma_{11}+\eI_{DP})
 ^{-1})^{1/2} }  \mathcal{Q}(dQ| \by_{1:P}) 
\end{equation}
with 
\[
 \mathcal{Q}(dQ | \by_{1:P}) =
 \frac{e^{-\frac{1}{2} \Psi (Q|\by_{1:P}) }\mathcal{Q}(dQ)  }{\int_{\CS^+_{D}}e^{-\frac{1}{2} \Psi (Q| \by_{1:P}) } \mathcal{Q}(dQ) } . 
\]
Note that  $\mathcal{Q}(dQ| \by_{1:P})=\mu(dQ| \by_{1:P})$. 
Define now
\[
\begin{split}
\bs_{0:P} = 
\begin{pmatrix}
\bs_{0} \\
 \bs_{1:P}  \\
\end{pmatrix}
=
 \begin{pmatrix}
   \text{M}_{0|1}  \bz_{1:P}   + \Sigma_{0|1}^{1/2} \bz_0 \\ 
   \Sigma_{11}^{\frac{1}{2}} \bz_{1:P}  \\ 
\end{pmatrix}
= A  \bz_{0:P} 
\end{split}
\]
where
\[
\text{M}_{0|1}= \Sigma_{01} \Sigma_{11}^{-1}  \Sigma_{11}^{\frac{1}{2}}  \qquad \text{and} \qquad 
A =  \begin{pmatrix}
 \Sigma_{0|1}^{1/2} &   \text{M}_{0|1}   \\ 
\mathbf{0} &    \Sigma_{11}^{\frac{1}{2}}  \\ 
\end{pmatrix}
\]
and
\[
\Sigma_{0|1}:= \Sigma_{00}-\Sigma_{01} \Sigma_{11}^{-} \Sigma_{01}^\top. 
\]
 Recall that $\Sigma$ is a function of $Q$ and hence  $A=A(Q)$ and $M_{0|1}=M_{0|1}(Q)$. 
From the well-known 
conditional distribution of a normal vector,  see e.g. Proposition 3.13 in \cite{Eaton2007},  one has that (given $Q$)
\[
\CN (d\bs_0| \Sigma_{01} \Sigma_{11}^{-} \bs_{1:P}, \Sigma_{0|1 } ) 
\CN (d\bs_{1:P}| \mathbf{0}, \Sigma_{11})=\CN (d\bs_{0:P} |  \mathbf{0}, \Sigma ),
\]
which shows that the conditional distribution of $\bs_{0:P}$ given $Q$ is a Gaussian 
with mean  $\mathbf{0}$ and covariance matrix  $\Sigma(Q)$. 
At this stage using also \eqref{zQ|y}
\[
\begin{split}
 \mu(d\bs_{0:P} d\bz_{0:P},Q| \by_{1:P})& =
\mu (d\bs_{0:P}| Q, \bz_{0:p},\by_{1:P} ) \mu(d\bz_{0:P} d Q | \by_{1:P}) \\
& = \delta_{A(Q) \bz_{0:p}  }(d\bs_{0:P})\mu(d\bz_{0})  \mu(d\bz_{1:P}| Q,  \by_{1:P}) \mu(d Q | \by_{1:P}) \\
\end{split}
\]
Moreover,  from \eqref{cond_Z_givenQy} one has 
\[
\mu(d\bz_{0:P}| Q, \by_{1:P})=\CN(d\bz_0|\mathbf{0},\eI_{D}) \CN(d\bz_{1:P} | \mathbf{m}_1(Q|\by_{1:P}),  ( {\beta} \Sigma_{11}(Q)+\eI_{DP})^{-1}).
\]
Hence the  conditional distribution of $\bz_{0:P}$ given $(Q,  \by_{1:P})$ is a Gaussian  
vector with mean and covariance matrix 
\[
\mathbf{m}(Q|\by_{1:P})= \begin{pmatrix}
\mathbf{0}^\top \\ 
\mathbf{m}_1(Q|\by_{1:P}) \\
\end{pmatrix}
\quad \text{and} \quad 
B(Q): = \begin{pmatrix}
\eI_{D} & \mathbf{0}  \\
\mathbf{0} & ( {\beta} \Sigma_{11}(Q)+\eI_{DP})^{-1} \\
\end{pmatrix}.
\]
In conclusion, the conditional distribution of $\bs_{0:P}=A(Q) \bz_{0:P}$ is a Gaussian distribution 
with mean 
\[
\mathbf{m}^*(Q|\by_{1:P})=A(Q) \mathbf{m}(Q|\by_{1:P})
\quad \text{and} \quad  
\Sigma^*(Q)=A(Q)  B(Q)A^\top(Q) .
\]

At this stage,  from 
\eqref{Post_N-bayes1} 
it follows that 
\[
P_{N,\mathrm{post}}(d\bs_{0:P}| \bY, \tilde \bX)  = \mu(d\bs_{0:P}| \by_{1:P})
\]
for the special choice $\Sigma(Q)=\Sigma(Q|\tilde{\bX})$ in \eqref{def_sigmaQ} and 
$\mathcal{Q}=\mathcal{Q}_{L,N}(dQ)$.  Similarly one gets $P_{\infty,\mathrm{post}}$ 
taking  $\mathcal{Q}=\mathcal{Q}_{\infty}$. At this stage, simple algebraic computations give Proposition \ref{prop2A}.


\end{document}